%% file: main.tex
\documentclass{bmvc2k}

\usepackage{amsmath,amssymb,bm}
\usepackage{booktabs}
\usepackage{colortbl}
\usepackage{multirow}
\usepackage{xcolor}
\usepackage{pifont}
\usepackage{microtype}
\usepackage{placeins}
\usepackage{dblfloatfix}
\usepackage{url}
\title{LiDSeg: Linear DehazeFormer Guidance for Efficient Semantic Segmentation in Adverse Weather Conditions}

\newcommand{\ucy}{
University of Cyprus\\
Nicosia, Cyprus
}

\addauthor{Yobsan Bayisa Leta}{}{1,2}
\addauthor{Christos Kyrkou}{}{1}
\addauthor{Maria K. Michael}{}{1,2}
\addauthor{Theocharis Theocharides}{}{1,2}

\addinstitution{
KIOS Research and Innovation Center of Excellence\\
\ucy
}

\addinstitution{
Department of Electrical and Computer Engineering\\
\ucy
}

\runninghead{Leta et al.}{LiDSeg}

\definecolor{lightgray}{gray}{0.93}
\newcommand{\ours}[1]{\textbf{#1}}
\newcommand{\second}[1]{\underline{#1}}
\newcommand{\NA}{--}

\newcommand{\diag}[1]{\textit{#1}}

\hypersetup{
  pdftitle={LiDSeg: Linear DehazeFormer Guidance for Efficient Semantic Segmentation in Adverse Weather Conditions},
  pdfauthor={Yobsan Bayisa Leta, Christos Kyrkou, Maria K. Michael, Theocharis Theocharides},
  pdfsubject={Semantic segmentation in adverse weather},
  pdfkeywords={semantic segmentation, adverse weather, real-time inference, LiDSeg}
}

\begin{document}

\maketitle

\begin{abstract}
Robust semantic segmentation in adverse weather conditions is essential for safety-critical autonomous systems. Current approaches generally follow two paradigms: sequential restore-then-segment pipelines, which introduce significant latency and prioritize pixel-level restoration over semantic discriminability, and domain adaptation techniques, which often struggle with representation bottlenecks and computational complexity. To tackle these challenges, we introduce LiDSeg, a framework that leverages reconstruction-informed guidance to improve segmentation under adverse weather. LiDSeg adopts an asymmetric training strategy in which a lightweight guidance encoder is jointly optimized with the segmentation branch under a combined segmentation and auxiliary reconstruction objective, encouraging the learning of features that are robust to weather corruptions. At inference, the reconstruction decoder is discarded to reduce computational overhead, while the guidance encoder is retained to extract weather-robust semantic features. A lightweight gated fusion (LGF) module then dynamically integrates these guidance features into the main segmentation branch, preserving clear-weather accuracy while substantially improving adverse-weather robustness.
Experiments across multiple datasets show that LiDSeg achieves a favorable balance between robustness and computational efficiency. On Foggy Cityscapes, it limits the clear-to-dense-fog mIoU degradation to only $2.5$--$2.8$ percentage points, reducing the baseline performance drop by more than $5\times$. On the official ACDC test server, LiDSeg-PIDNet reaches $75.07\%$ pooled all-condition mIoU, comparable to the supervised HRNet-W48 result of $75.0\%$, while remaining substantially smaller. The gains also generalize in a zero-shot setting to Foggy Driving and Foggy Zurich, where LiDSeg uses no real-fog images for training. More importantly, this robustness is achieved through an adaptable, plug-and-play design that adds only $1.9$--$3.1$M parameters across several models and costs no inference-time reconstruction overhead. The resulting models remain highly compact, ranging from $5.2$M to $39.5$M total parameters, and deliver real-time inference on resource-constrained edge devices such as the NVIDIA Jetson AGX Orin. Code is available at \footnote{\url{https://github.com/yobsanb/LiDSeg}}.
\end{abstract}

\section{Introduction}

\begin{figure}[htbp]
    \centering
    \includegraphics[width=1.0\linewidth]{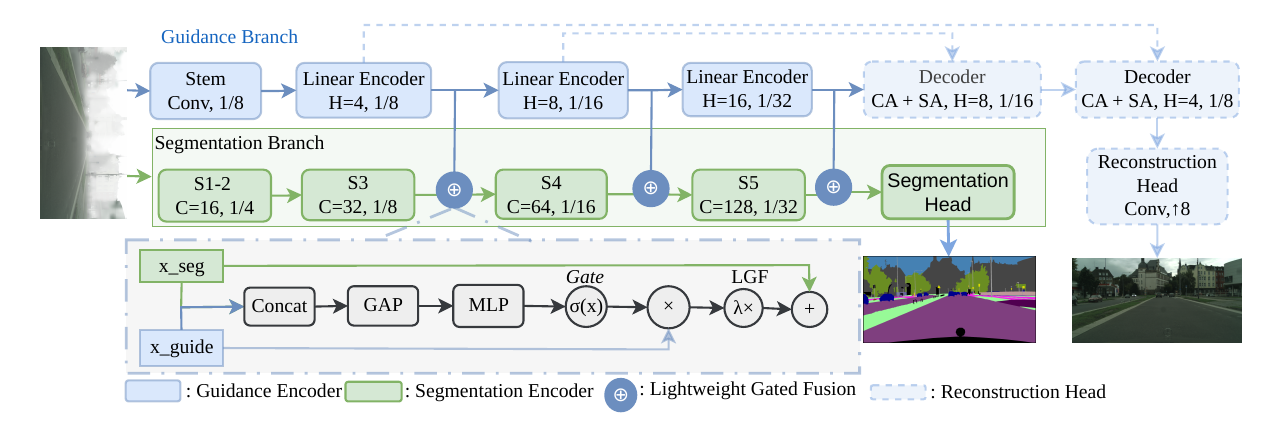}
    \caption{LiDSeg architecture. The degraded image $\mathbf{I}$ is processed by a guidance branch (Linear DehazeFormer encoder with training-only decoder) and a segmentation branch.}
    \label{fig:LiDSeg_detail}
\end{figure}

Semantic segmentation in adverse weather conditions is a fundamental challenge for diverse autonomous systems and safety-critical applications \cite{pan2025exploring}. Despite significant advances in clear-weather datasets, such as Cityscapes \cite{csurka2022semantic}, model performance declines in adverse weather conditions due to weather-induced artifacts that corrupt the semantic features essential for scene understanding \cite{yang2024semantic, zhang2023weatherstream, bi2024generalized}. Common methods for addressing this problem include sequential restore-then-segment pipelines that explicitly restore degraded images before segmentation \cite{wang2024two, niu2020effective}. While improving the visual quality of the restored image, these methods introduce additional inference latency and emphasize pixel-level restoration objectives, which do not necessarily enhance semantic discriminability~\cite{sun2024d2sl}. Another method is leveraging domain-adaptation techniques to align feature distributions across weather domains \cite{tsai2018learning, ganin2016domain, sakaridis2021curriculum, lee2022fifo, pan2025exploring, huang2022spc, cumformer2024}. However, this method often leads to representation bottlenecks and limits the robustness gains \cite{cumformer2024}.
To tackle these challenges, LiDSeg employs a modular framework with asymmetric training and inference that integrates weather-robust guidance into a segmentation network (Fig.~\ref{fig:LiDSeg_detail}).
During training, the auxiliary reconstruction loss, computed on paired adverse and clear images, jointly optimizes the shared guidance encoder together with the segmentation loss, so the encoder learns features robust to the distribution of weather corruptions. During inference, the reconstruction decoder is removed for efficiency, whereas the LGF module selectively fuses the encoder features into the segmentation network to enhance robustness in adverse weather conditions without compromising clear-weather accuracy. This design augments a standard segmentation network with minimal additional parameters and FLOPs, thereby avoiding the computational burden and representation conflicts of full-image restoration. Therefore, the objective of LiDSeg is to augment conventional real-time semantic segmentation networks that are optimized on clean-weather accuracy for weather robustness. The main contributions of this work are as follows.

\begin{enumerate}
\item We introduce LiDSeg, an end-to-end framework that augments a real-time semantic segmentation network with a lightweight linear-attention guidance to improve robustness in adverse weather conditions, without full-image reconstruction at inference.
\item We utilize an asymmetric training strategy in which auxiliary image reconstruction supervises the guidance encoder to learn features robust to weather corruptions, and the reconstruction decoder is discarded at inference to preserve deployment efficiency while retaining weather-robust guidance.
\item We design a Lightweight Gated Fusion (LGF) module that selectively integrates multiscale guidance into the semantic segmentation networks, preserving clear-weather accuracy while enhancing robustness in fog, rain, snow, and night.
\end{enumerate}
LiDSeg was evaluated on multiple benchmarks. On Foggy Cityscapes, guided variants reach $74.1{-}74.3\%$ dense-fog mIoU within $0.5{-}0.7$ below FreD-ViT ($109$M) at $9{-}21{\times}$ fewer parameters. On ACDC, LiDSeg shows similar robustness across several adverse weather conditions.
\section{Related Works}\label{Related}
Balancing segmentation accuracy and efficiency is critical for resource-constrained devices. Although multibranch architectures, such as BiSeNetV2~\cite{bisenetv2}, PIDNet~\cite{pidnet}, DDRNet~\cite{hong2021ddrnet}, STDC~\cite{fan2021stdc}, and transformer-based approaches, such as TopFormer~\cite{topformer},  RTLinearFormer~\cite{rtlinearformer2025}, and SeaFormer \cite{wan2023seaformer}, achieve efficient feature extraction on clear-weather datasets, they lack explicit mechanisms to handle feature degradation~\cite{csurka2022semantic} in adverse weather. The direct application of these standard segmentation models to adverse conditions leads to significant mIoU drops~\cite{nekrasov2018light,bisenetv2}. For instance, the BiSeNetV2 mIoU falls from $69.1\%$ on clear Cityscapes to $54.2\%$ on dense foggy Cityscapes ($\beta{=}0.02$), and the DDRNet mIoU falls from $77.8\%$ to $56.1\%$. Similar degradation occurs in ACDC and zero-shot datasets. Even robust conventional semantic networks, such as DeepLabV3+~\cite{chen2018deeplabv3plus} and HRNet~\cite{wang2020deep} struggle despite their large model size~\cite{zhang2023weatherstream, bi2024generalized}. On the other hand, although large transformers, such as Mask2Former~\cite{cheng2022} enhance robustness, their expensive computational requirements restrict their applicability to resource-constrained devices. Recent weather-aware aggregation~\cite{pan2025exploring} and language-guided approaches~\cite{wang2024two} have also added complexity without fully resolving domain shifts. Overall, to address the challenge of semantic segmentation in adverse weather conditions, current methods typically fall into two main categories.

\textbf{Restore-then-Segment.}
Sequential pipelines, such as TransWeather~\cite{valanarasu2022transweather} and WeatherStream~\cite{zhang2023weatherstream} treat restoration as a preprocessing step. Models, such as Restormer~\cite{restormer}, DehazeFormer~\cite{dehazeformer}, and LDFormer~\cite{pushpalatha2025ldformer} employ attention mechanisms for high-fidelity restoration before segmentation. However, this approach introduces significant latency and prioritizes pixel fidelity over semantic robustness~\cite{bi2024generalized}. Restoration often results in over-smoothed artifacts that obscure fine details. In addition, the disjoint sequential optimization of restoration and segmentation causes error accumulation~\cite{sun2024d2sl, pushpalatha2025ldformer}. LiDSeg differs from sequential restoration-then-segment and restoration–segmentation multitasking by using dehazing only as lightweight training-time guidance, with no restoration at inference. It linearizes attention in the dehazing encoder and uses reconstruction solely to supervise the guidance branch, reducing computation and avoiding restoration-induced error accumulation. LGF then selectively integrates multiscale guidance features instead of using generic concatenation or U-Net-style fusion.

\textbf{Domain Adaptation.}
Alternatively, domain-adaptation techniques bridge the feature gap through style transfer~\cite{lee2022fifo, pan2025exploring, huang2022spc} and adversarial learning~\cite{vu2019advent, yang2020fda}. Frequency-based methods, such as FreD~\cite{bi2024generalized}, also attempt to separate style from structure and are trained using unsupervised domain adaptation (UDA). However, UDA approaches, such as CumFormer~\cite{cumformer2024}, often involve high training complexity for marginal improvements in accuracy. Furthermore, although recent frameworks that leverage temporal correlation~\cite{kou2025adverse} and generalizable systems~\cite{yang2024semantic} have improved adaptation, they remain limited by computational inefficiencies. On the other hand,
recent adverse-weather segmentation models, such as (HRDA~\cite{hoyer2022hrda}, MIC~\cite{hoyer2023mic}, HALO~\cite{han2024halo}, CoDA~\cite{liu2024coda}, WA$^{2}$Net~\cite{pan2025exploring}) rely on ${>}85$M backbones and report no edge latency, whereas real-time architectures (BiSeNetV2, DDRNet, STDC) lack explicit weather adaptation. LiDSeg co-designs reconstruction supervision and channel gating for image degradations in resource-constrained environments, where reconstruction is used as an auxiliary task to penalize weather-specific encodings, and LGF modulates guidance with minimal overhead.

\section{Proposed Methodology}
\subsection{LiDSeg Problem Formulation and Overview}

Adverse weather conditions degrade image quality and corrupt semantic features essential for accurate scene understanding. This degradation can be mathematically formalized as:
\begin{equation}
  \mathbf{I} = \mathcal{D}(\mathbf{J};\,\bm{\theta}),
  \label{eq:degradation}
\end{equation}
where $\mathbf{J}$ denotes the clean image, $\mathbf{I}$ represents the observed degraded image, and $\mathcal{D}(\cdot)$ is a degradation parameterized by condition-specific degradation variables $\bm{\theta}$ (encompassing rain, snow, fog, and nighttime illumination). Depending on the adverse condition, $\mathcal{D}$ exhibits distinct yet structurally related physical degradation, from sparse additive occlusions (e.g., rain streaks~\cite{garg2004vision}) to dense, depth-dependent multiplicative attenuation and global illumination changes~\cite{narasimhan2002vision, sakaridis2018semantic}. These conditions simultaneously degrade local high-frequency details and global contextual cues~\cite{li2019benchmarking}. Across fog, rain, and snow, much of the corruption is channel-coherent, which motivates channel-wise fusion rather than heavy spatial attention; night is the partial exception (Sec.~\ref{sec:acdc_perclass}), where this assumption weakens.

Because semantic annotations depend on static scene content, which remains unaffected by weather, the objective is to learn $f(\cdot)$ that is robust to weather-induced degradations:
\begin{equation}
f(\mathcal{D}(J;\theta)) \approx f(J), \quad \forall\theta .
  \label{eq:invariance}
\end{equation}
To enforce this robustness, LiDSeg introduces a dual-branch architecture (Fig.~\ref{fig:LiDSeg_detail}), where the segmentation branch uses lightweight models~\cite{bisenetv2,pidnet,hong2021ddrnet,fan2021stdc} and the guidance branch uses a linearized encoder inspired by DehazeFormer ~\cite{dehazeformer} to extract multi-scale features. LGF modules (Eq.~\eqref{eq:lgf}) integrate guidance into the segmentation branch with minimal overhead.
\subsection{Guidance Encoder Architecture}
The guidance encoder extracts multiscale features to support the robustness objective in Eq.~\eqref{eq:invariance}. It produces feature maps at resolutions $1/8$, $1/16$, and $1/32$ with channel dimensions $\{32, 64, 128\}$, matching the segmentation branch to enable direct fusion without additional projection or resampling layers.

Processing high resolution images using resource constrained devices requires efficient attention mechanisms. Standard self-attention uses $\mathrm{softmax}(QK^\top)V$ with $\mathcal{O}(N^2)$ complexity, making it difficult to process high-resolution features on resource-constrained devices. To address this, we adopt a linear-attention formulation inspired by~\cite{katharopoulos2020transformers}, which exploits matrix associativity to compute the key-value product:
\begin{equation}
    \mathrm{Attn}(\mathbf{Q}, \mathbf{K}, \mathbf{V})
    = \mathbf{Q}\left(\rho(\mathbf{K})^\top \mathbf{V}\right),
    \label{eq:lin_attn}
\end{equation}
where $\mathbf{Q}, \mathbf{K}, \mathbf{V} \in \mathbb{R}^{N \times D}$ are spatially flattened features, and $\rho(\cdot)$ denotes softmax along the spatial dimension. The key-value product $\rho(\mathbf{K})^\top\mathbf{V} \in \mathbb{R}^{D \times D}$ produces a compact $D \times D$ context matrix rather than an $N \times N$ attention matrix, reducing complexity to $\mathcal{O}(N \cdot D^2)$. To limit the propagation of weather artifacts through global feature aggregation, we add a learnable gate inspired by Gated Linear Units~\cite{dauphin2017language,so2021primer}:
\begin{equation}
    \mathrm{GLA}(\mathbf{x})
    = \left(
    \mathrm{Attn}(\mathbf{Q}, \mathbf{K}, \mathbf{V})
    \odot \phi(\mathbf{G})
    \right)\mathbf{W}_{O},
    \label{eq:gla}
\end{equation}
where $\mathbf{x} \in \mathbb{R}^{N \times D}$ denotes the input feature map and $\mathbf{W}_{O} \in \mathbb{R}^{D \times D}$ is the learnable output projection. The attention components are computed via linear transformations: $\mathbf{Q}=\mathbf{x}\mathbf{W}_Q$, $\mathbf{K}=\mathbf{x}\mathbf{W}_K$, and $\mathbf{V}=\mathbf{x}\mathbf{W}_V$. The gating signal $\mathbf{G} \in \mathbb{R}^{N \times D}$ is derived from the input as $\mathbf{G}=\mathbf{x}\mathbf{W}_G$, where $\mathbf{W}_Q, \mathbf{W}_K, \mathbf{W}_V, \mathbf{W}_G \in \mathbb{R}^{D \times D}$ are learned weights, and $\phi(\cdot)$ is the SiLU activation. This generates a soft mask that suppresses weather-artifact responses across channels through element-wise multiplication ($\odot$) before the features are projected by $\mathbf{W}_{O}$.

Although attention captures global context, it lacks explicit modeling of local spatial patterns, such as edges and fine textures~\cite{xie2021segformer}. To complement the global receptive field of GLA, each encoder block incorporates a Gated-Dconv Feed-Forward Network (GDFN)~\cite{restormer} with depth-wise convolutions for local spatial mixing:
\begin{align}
    \hat{\mathbf{x}}^{(l)}
    &= \mathbf{x}^{(l-1)}
    + \mathrm{GLA}\!\left(\mathrm{LN}(\mathbf{x}^{(l-1)})\right),
    \nonumber\\
    \mathbf{x}^{(l)}
    &= \hat{\mathbf{x}}^{(l)}
    + \mathrm{GDFN}\!\left(\mathrm{LN}(\hat{\mathbf{x}}^{(l)})\right),
    \label{eq:block}
\end{align}
\begin{equation}
    \left[\mathbf{x}_{1}, \mathbf{x}_{2}\right]
    = \mathrm{DWConv}\!\left(\mathbf{W}_{\mathrm{in}}(\mathbf{x})\right),
    \qquad
    \mathrm{GDFN}(\mathbf{x})
    = \mathbf{W}_{\mathrm{out}}
    \left(
    \mathrm{Gate}(\mathbf{x}_{1}) \odot \mathbf{x}_{2}
    \right),
    \label{eq:gdfn}
\end{equation}
where $\mathrm{LN}$ is Layer Normalization, $\mathrm{DWConv}$ is depth-wise convolution, $\mathrm{Gate}(\cdot)$ denotes the gating activation, and $\mathbf{W}_{\mathrm{in}}$ and $\mathbf{W}_{\mathrm{out}}$ are linear projections. Together, GLA and GDFN enable the encoder to learn features that are both globally coherent and locally precise, which is essential for robust segmentation under weather degradation.

\subsection{Auxiliary Reconstruction Decoder}

The guidance encoder learns weather-robust representations using an auxiliary reconstruction decoder. This decoder follows a U-Net-like architecture with Restormer-style refinement~\cite{restormer}. At each decoder stage $i \in \{2, 1\}$,
\begin{equation}
    \mathbf{d}_i = \Phi_i \left( \mathbf{W}_i^{\mathrm{reduce}} \left[ \mathrm{Up}(\mathbf{d}_{i+1}) \,\Vert\, \mathbf{e}_i \right] \right),
    \label{eq:decoder}
\end{equation}
where $\mathbf{d}_i, \mathbf{e}_i \in \mathbb{R}^{C_i \times H_i \times W_i}$ denote the decoder and encoder features at stage $i$ with channel dimension $C_i$, $\mathrm{Up}(\cdot)$ is PixelShuffle with scale factor 2, $[\cdot \,\Vert\, \cdot]$ denotes channel-wise concatenation, $\mathbf{W}_i^{\mathrm{reduce}} \in \mathbb{R}^{C_i \times 2C_i}$ is a $1 \times 1$ convolution reducing channel from $2C_i$ to $C_i$, and $\Phi_i(\cdot)$ denotes a stack of Restormer-like blocks (channel-wise attention and GDFN ) for feature refinement. The decoder output passes through a progressive upsample that applies three successive $2\times$ PixelShuffle operations (from $1/8$ to full resolution) with intermediate refinement. The final reconstruction employs global residual learning as follows:
\begin{equation}
    \mathbf{x}_{\mathrm{rec}} = \mathbf{x}_{\mathrm{in}} + \Delta \mathbf{x},
\end{equation}
where $\mathbf{x}_{\mathrm{in}} \in \mathbb{R}^{3 \times H \times W}$ is the degraded image and $\Delta \mathbf{x} \in \mathbb{R}^{3 \times H \times W}$ is the residual predicted by the decoder. The auxiliary decoder adds ${\sim}15\%$ peak GPU memory during training only.

\subsection{Lightweight Gated Fusion (LGF)}
Given the spatial and condition dependent nonuniformity of weather degradation, we propose an LGF that selectively fuses guidance features into segmentation features at multiple stages using a channel-wise gating mechanism. At fusion stages $k \in \{3,4,5\}$, the fused features are computed as

\begin{equation}
    \mathbf{f}_k' = \mathbf{f}_k + \lambda_k \cdot \sigma\!\left(\mathbf{W}_2 \, \delta\!\left(\mathbf{W}_1 \, \text{GAP}([\mathbf{f}_k, \mathbf{g}_k])\right)\right) \odot \mathbf{g}_k,
    \label{eq:lgf}
\end{equation}
where $\mathbf{f}_k, \mathbf{g}_k \in \mathbb{R}^{C_k \times H_k \times W_k}$ denote semantic and guidance features ($C_k \in \{32,64,128\}$), respectively. Leveraging the squeeze-and-excitation principle~\cite{hu2020squeeze}, we concatenate the inputs and apply Global Average Pooling (GAP) to extract global context, followed by a bottleneck MLP (reduction $r{=}8$, weights $\mathbf{W}_{1,2}$) with ReLU $\delta$ and sigmoid $\sigma$ to generate channel-wise gates. Here $\lambda_k$ is a learnable per-stage scalar that controls how much guidance is needed at each scale; the ``$\lambda{\times}$'' in Fig.~\ref{fig:LiDSeg_detail} denotes this learned gate. Weather corruption in fog, rain, and snow is predominantly channel-coherent (Eq.~\eqref{eq:degradation}); we verify this by measuring the cross-channel correlation of the clear$\to$degraded residual, which is high for fog/rain/snow ($\bar{\rho}{>}0.8$) but lower for night ($\bar{\rho}{\approx}0.5$, consistent with Sec.~\ref{sec:acdc_perclass}). Channel gating targets this structure with far fewer parameters and FLOPs, whereas spatial or cross-attention fusion costs more fusion parameters for only a marginal accuracy gain (Tab.~\ref{tab:architecture_ablation}). These gates dynamically modulate $\mathbf{g}_k$, preserving clear-weather accuracy while enhancing robustness in adverse conditions.

\subsection{Loss Function}
\label{sec:loss}

The training loss combines segmentation and reconstruction to achieve a balance between the segmentation focus on discriminative features and the reconstruction need for pixel-level accuracy. Therefore, the total loss $\mathcal{L}_{\text{total}}$ is computed using the task-dependent uncertainty weight \cite{kendall2018multi} as shown in Eq.~\ref{eq:total_loss}.
\begin{equation}
\mathcal{L}_{\text{total}} = \frac{1}{\sigma_{\text{seg}}^2}\mathcal{L}_{\text{seg}} + \frac{1}{\sigma_{\text{rec}}^2}\mathcal{L}_{\text{rec}} + \log \sigma_{\text{seg}}^2 + \log \sigma_{\text{rec}}^2,
\label{eq:total_loss}
\end{equation}
where $\sigma_{\text{seg}}^2$ and $\sigma_{\text{rec}}^2$ are learnable task uncertainties. Higher uncertainty downweights the corresponding loss, whereas the log regularizes to prevent the trivial solution $\sigma \to \infty$.

\textbf{Segmentation loss.}
$\mathcal{L}_{\text{seg}}$ combines cross-entropy, Lov\'asz-Softmax~\cite{berman2018lovasz}, and auxiliary supervision:
\begin{equation}
\mathcal{L}_{\text{seg}} = \mathcal{L}_{\text{CE}} + \lambda_L \mathcal{L}_{\text{Lov}} + \lambda_{\text{aux}} \mathcal{L}_{\text{aux}}.
\end{equation}
We set $\lambda_L = 0.5$ and $\lambda_{\text{aux}} = 0.4$. These values follow established practice: $\lambda_{\text{aux}} = 0.4$ matches the deep-supervision weighting in PSPNet~\cite{zhao2017pyramid} and BiSeNetV2~\cite{bisenetv2}, ensuring that auxiliary heads do not dominate gradients. The weight $\lambda_L = 0.5$ balances the IoU-based Lovász optimization against pixel-wise cross-entropy, as recommended in~\cite{berman2018lovasz}. A sensitivity sweep (in supplementary) shows LiDSeg is robust to this choice.

\textbf{Reconstruction loss.}
For reconstruction, the Charbonnier loss is used for training.
\begin{equation}
    \mathcal{L}_{\text{rec}} = \frac{1}{N} \sum_{n=1}^{N} \sqrt{\left( \hat{\mathbf{x}}_n - \mathbf{x}_n^{\mathrm{clear}} \right)^2 + \epsilon^2}, \quad \epsilon = 10^{-3},
    \label{eq:lrec}
\end{equation}
where $N$ is the number of pixels, $\hat{\mathbf{x}}_n$ is the $n$-th pixel of the reconstructed image, $\mathbf{x}_n^{\mathrm{clear}}$ is the clear ground-truth pixel, $\epsilon $ is used for numerical stability. The adverse-to-clear mapping is not assumed to be uniquely identifiable because weather can obscure scene information~\cite{nayar1999vision}. LiDSeg therefore uses $\mathcal{L}_{\mathrm{rec}}$ only as a pretraining regularizer for degradation-stable guidance features rather than requiring a unique photorealistic reconstruction at deployment; the decoder is discarded afterward, and Sec.~\ref{sec:ablation} isolates the contribution of this supervision.
\section{Experimental Results and Discussion}\label{sec:experiments}
\subsection{Experimental Setup}
We implemented LiDSeg in PyTorch and trained on NVIDIA Tesla V100 GPUs using the AdamW optimizer ($\mathrm{lr}{=}10^{-3}$, weight decay $10^{-4}$), automatic mixed precision, and a cosine learning rate schedule. During paired Foggy Cityscapes pretraining, the guidance encoder is optimized with both segmentation and reconstruction objectives. For ACDC fine-tuning, the reconstruction decoder and $\mathcal{L}_{\mathrm{rec}}$ are removed, and the segmentation branch, LGF, and pretrained guidance encoder are fine-tuned using segmentation labels only. We evaluate both \textit{Uber} configurations (optimized across all adverse weather conditions simultaneously) and condition-specific Expert models (trained on a single weather condition each). Inference latency is measured in frames per second (FPS) on an NVIDIA Jetson AGX Orin.\\
\noindent\textbf{Datasets.} In reality, capturing perfectly aligned pairs of clear and adverse weather image for training is nearly impossible because the same dynamic scene (with moving objects, vehicles, and pedestrians) cannot be recorded at the exact same timestamp under both clear and adverse weather without any other variations, making reliable ground-truth annotation extremely difficult. Therefore, we rely on synthetic pairs that provide pixel-level alignment between clean and weather-affected images while isolating only the weather effects to learn robust features. Given this reality, we pretrain on Foggy Cityscapes~\cite{sakaridis2018semantic}, which provides synthetic paired clear and foggy images across a progressive curriculum of attenuation coefficients $\beta \in \{0.005, 0.01, 0.02\}$. This also allows direct comparison with recent works on adverse weather conditions such as FreD~\cite{bi2024generalized}, IBNet~\cite{pan2018ibn}, and IterNorm~\cite{huang2019iterative}. Real-world evaluation and target-domain fine-tuning (which requires no clear image pairs) utilize ACDC~\cite{sakaridis2021acdc}, spanning real fog, night, rain, and snow. Moreover, for evaluating generalization, we assessed guided models via zero-shot on Foggy Zurich v2~\cite{sakaridis2021curriculum} and Foggy Driving~\cite{sakaridis2018semantic}.\\
\noindent\textbf{Metrics.} We report mIoU per class and per condition. For ACDC, \textbf{Pooled All} denotes the benchmark's class-mean IoU over the pooled all-condition set and is not the arithmetic mean of Fog, Night, Rain, and Snow. To analyse the correlation between reconstruction and segmentation, we also track Peak Signal-to-Noise Ratio (PSNR) and Structural Similarity Index Measure (SSIM).
\subsection{Learned Feature Robustness and Training Dynamics}
\begin{figure}[b]
    \centering
    \includegraphics[width=\linewidth]{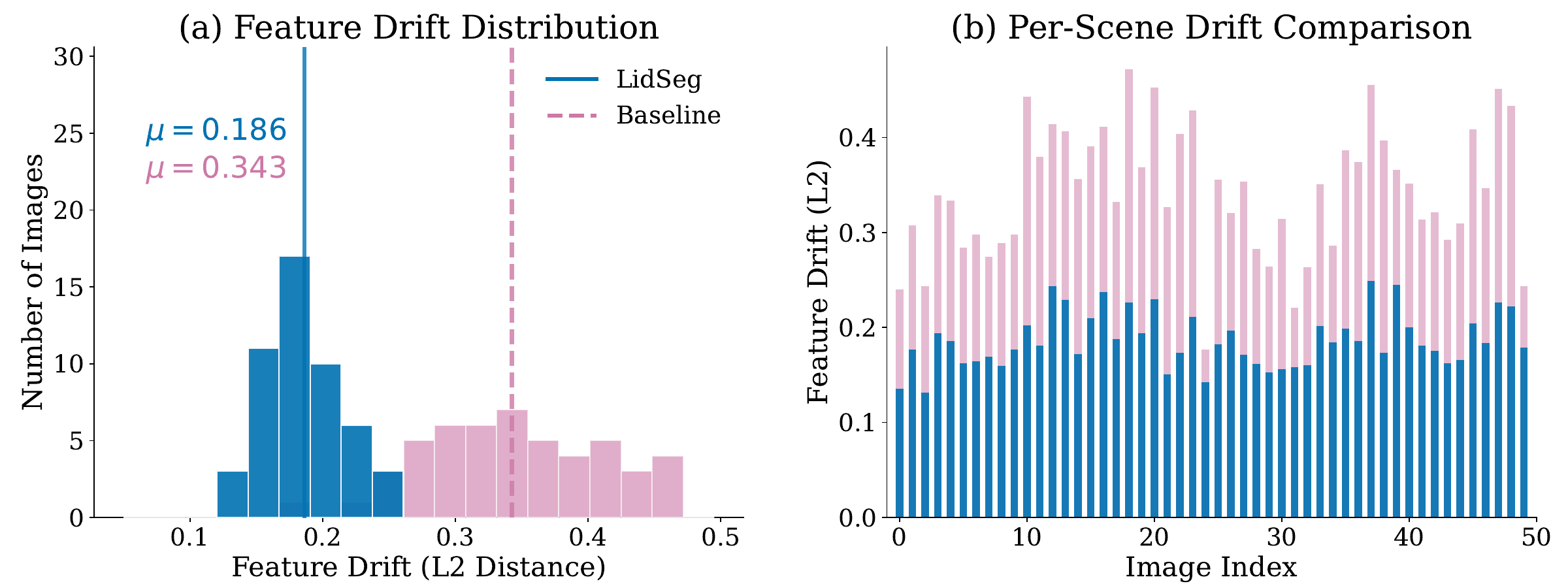}
    \caption{Feature stability on Foggy Cityscapes \emph{val} (BiSeNetV2) ($\beta{=}0.02$): (a) $L_2$ feature drift (LiDSeg $\mu{=}0.186$ vs.\ baseline $\mu{=}0.343$). (b) Per-scene drift (LiDSeg lower on all scenes).}
    \label{fig:feature_invariance}
\end{figure}

To evaluate model robustness, we analyze feature stability by quantifying the $L_2$ feature drift between corresponding clear and foggy scenes. As illustrated in the Feature Drift Distribution (Fig.~\ref{fig:feature_invariance}a), LiDSeg (blue) exhibits a narrow distribution clustered near zero, demonstrating better stability against weather-induced domain shifts compared to the baseline (pink). Quantitatively, LiDSeg substantially reduces the mean feature drift to $\mu=0.186$ relative to the baseline ($\mu=0.343$). This consistency is validated by per-scene comparisons (Fig.~\ref{fig:feature_invariance}b), where LiDSeg yields uniformly lower drift across all test scenes.

\subsection{Performance and stability on Foggy Cityscapes}
\begin{figure}[htbp]
    \centering
    \includegraphics[width=\linewidth]{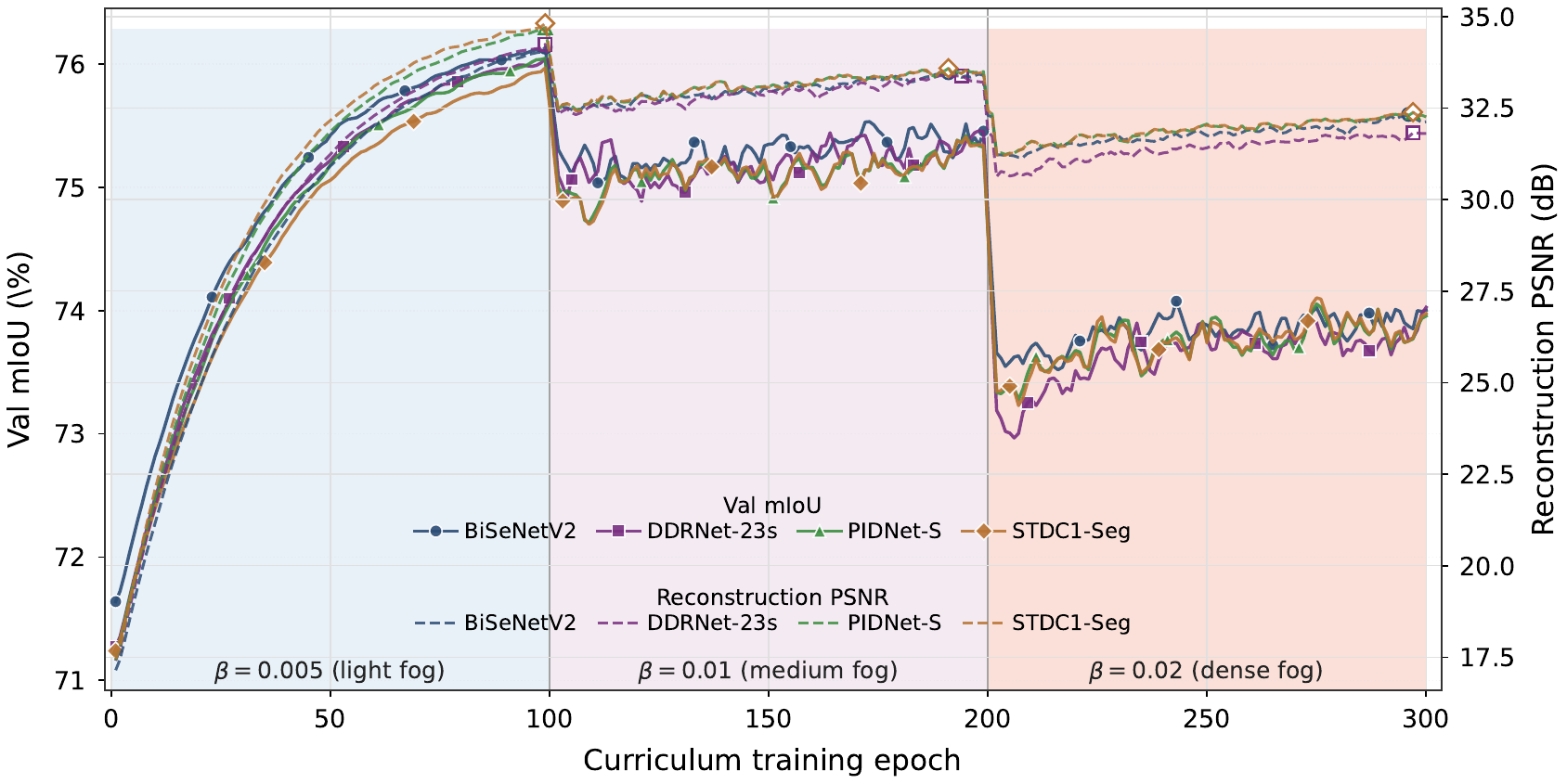}
    \caption{\textbf{Reconstruction and segmentation training dynamics on Foggy
    Cityscapes}  validation set (four guided models; fog curriculum
    $\beta{=}0.005{\to}0.01{\to}0.02$). Validation mIoU (\emph{solid}, left axis) and guidance-encoder reconstruction PSNR (\emph{dashed}, right axis).}
    \label{fig:training_dynamics}
\end{figure}

Fig.~\ref{fig:training_dynamics} plots validation mIoU and guidance-encoder reconstruction PSNR for all four guided models over the fog curriculum ($\beta{=}0.005{\to}0.01{\to}0.02$). Reconstruction quality and segmentation accuracy co-evolve throughout training, where both metrics slightly dip at each denser fog in a synchronized manner. At each transition, the encoder encounters a sudden distribution shift because denser fog attenuates high-frequency content, reducing reconstruction quality (lower PSNR) and perturbing the segmentation feature space (lower mIoU). However this shift is relatively small compared to baselines as illustrated in Tab. \ref{tab:sensitivity_foggy_density}. Additionally, the subsequent rise in both curves is consistent with the recovery of latent structure useful for segmentation. Those patterns are nearly identical across all LiDSeg variants despite architectural differences (bilateral fusion, dual-resolution, PID, STDC).
We also compared LiDSeg-guided models with adverse-weather and baseline models. Tab. ~\ref{tab:sensitivity_foggy_density} reports the Foggy Cityscapes validation mIoU at $\beta{\in}\{0.005,0.01,0.02\}$. LiDSeg rows are taken from the per-stage training of Fig.~\ref{fig:training_dynamics}. Every guided model degrades by only $2.5{-}2.8$ mIoU points from clear to dense fog. This drop is less than half that of FreD-ViT and substantially
smaller than those of the unguided baselines.  The scaled unguided models (BiSeNetV2-wide and DDRNet-23s-wide), widened to $5.2$/$7.7$M budget and trained on the identical curriculum without the guidance branch, reach only $56.9$/$57.8\%$ dense-fog mIoU. Therefore, scaling the model to match the parameter and FLOP budget recovers at most ${+}2.7$ mIoU points over the vanilla model, whereas LiDSeg recovers ${+}20.0$ points. This gain results from the reconstruction-guided robustness rather than added capacity. On dense-fog accuracy LiDSeg reaches $74.1{-}74.3\%$, within $0.5{-}0.7$ mIoU points of FreD-ViT ($74.8\%$) while using substantially fewer parameters, placing it on the robustness and efficiency frontier of adverse weather condition semantic segmentation design.
\begin{table*}[htbp]
  \centering
  \caption{Fog density robustness on Foggy Cityscapes \emph{val} (mIoU \%).
  \textbf{Clear} is clear-weather Cityscapes mIoU; $\boldsymbol{\Delta}$ is the clear$\to$dense ($\beta{=}0.02$) mIoU drop (smaller is better).
  GFLOPs at $1024{\times}2048$. $\dagger$: denotes incremental widening to match the guided parameter/FLOP budget and trained on an identical fog curriculum without the guidance branch.}
  \label{tab:sensitivity_foggy_density}
  \footnotesize
  \setlength{\tabcolsep}{3pt}
  \resizebox{\textwidth}{!}{
  \begin{tabular}{@{}l l c c c c c c c@{}}
    \toprule
    \textbf{Method} & \textbf{Backbone} & \textbf{Params} & \textbf{GFLOPs} &
      \textbf{Clear (mIoU)} & \textbf{Light (mIoU)} & \textbf{Medium (mIoU)} &
      \textbf{Dense (mIoU)} & $\boldsymbol{\Delta}$ \\
    \midrule
    \multicolumn{9}{@{}l}{\textit{Real-time semantic segmentation models}} \\
    BiSeNetV2~\cite{bisenetv2}              & \NA  & \textbf{3.36M} & 99.7  &
      69.1 & 60.9 & 58.8 & 54.2 & \textcolor{red}{$\downarrow$14.9} \\
    DDRNet-23s~\cite{hong2021ddrnet}        & \NA     & 5.70M  & \textbf{36.7}  &
      77.8 & 66.1 & 62.2 & 56.1 & \textcolor{red}{$\downarrow$21.7} \\
    STDC1-Seg~\cite{fan2021stdc}            & \NA       & 8.43M  & 166.3  &
      71.8 & 67.5 & 62.5 & 56.1 & \textcolor{red}{$\downarrow$15.7} \\
    PIDNet-L~\cite{pidnet}                  & \NA     & 37.31M  & 276.2  &
      74.0 & 68.4 & 63.0 & 58.9 & \textcolor{red}{$\downarrow$15.1} \\
    \midrule
    \multicolumn{9}{@{}l}{\textit{Unguided models with incremental (matched budget, same curriculum)}} \\
    BiSeNetV2-wide$^\dagger$                & BiSeNetV2  & 5.22M  & 108.8 &
      70.4 & 64.2 & 60.6 & 56.9 & \textcolor{red}{$\downarrow$13.5} \\
    DDRNet-23s-wide$^\dagger$               & DDRNet     & 7.65M  & 46.0 &
      78.2 & 68.5 & 64.0 & 57.8 & \textcolor{red}{$\downarrow$20.4} \\
    \midrule
    \multicolumn{9}{@{}l}{\textit{Models use heavy backbones}} \\
    IBNet~\cite{pan2018ibn}                 & ResNet-50  & ${>}$25.6M & ${>}$100 &
      76.5 & 72.4 & 67.9 & 59.5 & \textcolor{red}{$\downarrow$17.0} \\
    IterNorm~\cite{huang2019iterative}      & ResNet-50  & ${>}$25.6M & ${>}$100 &
      76.0 & 72.0 & 68.3 & 60.7 & \textcolor{red}{$\downarrow$15.3} \\
    SW~\cite{pan2019switchable}             & ResNet-50  & ${>}$25.6M & ${>}$100 &
      77.0 & 73.3 & 69.4 & 61.7 & \textcolor{red}{$\downarrow$15.3} \\
    ISW~\cite{choi2021robustnet}            & ResNet-50  & ${>}$25.6M & ${>}$100 &
      76.2 & 72.1 & 67.9 & 60.1 & \textcolor{red}{$\downarrow$16.1} \\
    DeepLabV3+~\cite{chen2018deeplabv3plus} & ResNet-101 & 62.7M & 2032 &
      79.6 & 67.1 & 65.2 & 61.6 & \textcolor{red}{$\downarrow$18.0} \\
    SegFormer~\cite{xie2021segformer}       & MiT-B4     & 64.1M & 1240 &
      81.0 & 70.5 & 66.1 & 62.3 & \textcolor{red}{$\downarrow$18.7} \\
    Mask2Former~\cite{cheng2022}            & Swin-B     & 107M  &  466 &
      83.5 & 76.2 & 74.5 & 70.8 & \textcolor{red}{$\downarrow$12.7} \\
    FreD-ViT~\cite{bi2024generalized}       & Swin-B     & 109M  &  466 &
      82.8 & \textbf{79.5} & \textbf{78.0} & \textbf{74.8} & \textcolor{red}{$\downarrow$8.0} \\
    \midrule
    \rowcolor{lightgray}
    \ours{LiDSeg (BiSeNetV2)} & \NA & 5.22M  & 108.8 &
      76.7 & \underline{76.3} & \underline{75.7} & 74.2 & \textcolor{red}{$\downarrow$2.5} \\
    \rowcolor{lightgray}
    \ours{LiDSeg (DDRNet-23s)} & \NA  & 7.65M  & 46.0 &
      76.9 & 76.2 & 75.6 & 74.1 & \textcolor{red}{$\downarrow$2.8} \\
    \rowcolor{lightgray}
    \ours{LiDSeg (STDC1-Seg)} & \NA    & 11.48M & 176.3 &
      76.9 & 76.2 & 75.5 & \underline{74.3} & \textbf{\textcolor{red}{$\downarrow$2.6}} \\
    \rowcolor{lightgray}
    \ours{LiDSeg (PIDNet-L)}  & \NA  & 39.54M & 285.9 &
      76.8 & 76.2 & 75.5 & 74.2 & \underline{\textcolor{red}{$\downarrow$2.6}} \\
    \bottomrule
  \end{tabular}
  }
\end{table*}
\subsection{LiDSeg Guidance Performance on ACDC}
\label{sec:acdc_test}
The objective of LiDSeg is to learn weather-robust features from synthetic paired images and transfer them to real-world datasets such as ACDC. During ACDC fine-tuning, no paired clear target is used: the reconstruction decoder and $\mathcal{L}_{\mathrm{rec}}$ are removed, and the segmentation branch, LGF, and guidance encoder are updated with ACDC segmentation labels only. Because supervised fine-tuning and UDA use different target-domain supervision, Tab.~\ref{tab:acdc_test} labels each protocol explicitly; UDA rows are included as task-level references for comparisons. LiDSeg-PIDNet obtains $75.07\%$ mIoU, comparable with supervised HRNet-W48 ($75.0\%$ mIoU, 696 GFLOPs) and other heavy UDA models, while night remains the most challenging condition.
\begin{table*}[htbp]
  \centering
  \caption{ACDC \emph{test} mIoU (\%). \textbf{Pooled All} is the benchmark pooled all-condition evaluation; \textbf{Mean} is the arithmetic mean of Fog, Night, Rain, and Snow where per-condition scores are available. Sup.: supervised ACDC labels; Sup.-FT: LiDSeg fine-tuned with ACDC segmentation labels after synthetic paired pretraining; UDA: adaptation with unlabeled target images. (\NA): not reported.}
  \label{tab:acdc_test}
  \footnotesize
  \setlength{\tabcolsep}{4.2pt}
  \resizebox{\textwidth}{!}{
  \begin{tabular}{@{}l l c c c c c c c c@{}}
    \toprule
    \textbf{Method} & \textbf{Protocol} & \textbf{Params} & \textbf{GFLOPs} &
      \textbf{Fog} & \textbf{Night} & \textbf{Rain} & \textbf{Snow} & \textbf{Pooled All} & \textbf{Mean} \\
    \midrule
    \multicolumn{10}{@{}l}{\textit{Supervised ACDC labels}} \\
    RefineNet~\cite{lin2017}                & Sup. & 118M      & \NA       & 65.7 & 55.5 & 68.7 & 65.9 & 65.3 & 64.0 \\
    DeepLabV2~\cite{sakaridis2021acdc}      & Sup. & 44M       & \NA       & 54.5 & 45.3 & 59.3 & 57.1 & 55.3 & 54.1 \\
    DeepLabV3+~\cite{chen2018deeplabv3plus} & Sup. & 62.7M     & 2032      & 69.1 & 60.9 & 74.1 & 69.6 & 70.0 & 68.4 \\
    HRNet-W48~\cite{wang2020deep}           & Sup. & 65.9M     & 696       & 74.7 & \textbf{65.3} & 77.7 & 76.3 & 75.0 & 73.5 \\
    Mask2Former~\cite{cheng2022}            & Sup. & 107M      & 466       & \NA & \NA & \NA & \NA & 77.3 & \NA \\
    ViT-Adapter~\cite{chen2022vitadapter}   & Sup. & ${>}$320M & ${>}$1000 & \NA & \NA & \NA & \NA & 78.4 & \NA \\
    \midrule
    \multicolumn{10}{@{}l}{\textit{UDA task-level references}} \\
    DANNet~\cite{wu2021dannet}              & UDA & \NA       & \NA       & \NA & \NA & \NA & \NA & 50.0 & \NA \\
    DAFormer~\cite{hoyer2022daformer}       & UDA & 84.0M     & \NA       & \NA & \NA & \NA & \NA & 55.4 & \NA \\
    HRDA~\cite{hoyer2022hrda}               & UDA & 85.7M     & 251       & \NA & \NA & \NA & \NA & 68.0 & \NA \\
    MIC~\cite{hoyer2023mic}                 & UDA & 85.7M     & 251       & \NA & \NA & \NA & \NA & 70.4 & \NA \\
    HALO~\cite{han2024halo}                 & UDA & 86.0M     & 259       & \NA & \NA & \NA & \NA & 71.9 & \NA \\
    CoDA~\cite{liu2024coda}                 & UDA & 90.5M     & 271       & \NA & \NA & \NA & \NA & 72.6 & \NA \\
    WA$^{2}$Net~\cite{pan2025exploring}     & UDA & ${>}$85M  & ${>}$255  & \NA & \NA & \NA & \NA & 73.8 & \NA \\
    \midrule
    \rowcolor{lightgray}
    \ours{LiDSeg (PIDNet-L)} & \textbf{Sup.-FT} & \textbf{39.54M} & \textbf{285.9} &
      73.99 & 64.58 & \textbf{78.58} & \textbf{77.18} & \textbf{75.07} & \textbf{73.58} \\
    \bottomrule
  \end{tabular}
  }
\end{table*}

\subsection{ACDC: Per-Condition Per-Class performance}
\label{sec:acdc_perclass}
Across all conditions, Uber models consistently outperform condition-specific Experts, showing that joint training across adverse conditions strengthens the shared representation. Cross-condition evaluation in Tab.~\ref{tab:supp_cross_condition} further supports this finding, with strong transfer among the channel-coherent fog, rain, and snow conditions. Tab.~\ref{tab:supp_acdc_perclass} reports the corresponding class-wise IoU results. Fog remains the strongest setting, consistent with Foggy-Cityscapes pretraining, whereas night remains the most challenging condition.
\begin{table*}[htbp]
  \centering
  \caption{All-conditions ACDC \emph{test} per-class IoU (\%). Comparison rows are reported by WA$^{2}$Net~\cite{pan2025exploring}. LiDSeg (PIDNet-L) is Uber training on ACDC. Bold and underlined values denote the best and second-best results, respectively.}
  \label{tab:acdc_all}
  \footnotesize
  \setlength{\tabcolsep}{1.8pt}
  \resizebox{\textwidth}{!}{
  \begin{tabular}{l|ccccccccccccccccccc|c}
    \toprule
    \textbf{Method} & \rotatebox{90}{\scriptsize Road} & \rotatebox{90}{\scriptsize Side.} & \rotatebox{90}{\scriptsize Buil.} & \rotatebox{90}{\scriptsize Wall} & \rotatebox{90}{\scriptsize Fenc.} & \rotatebox{90}{\scriptsize Pole} & \rotatebox{90}{\scriptsize Ligh.} & \rotatebox{90}{\scriptsize Sign} & \rotatebox{90}{\scriptsize Vege.} & \rotatebox{90}{\scriptsize Terr.} & \rotatebox{90}{\scriptsize Sky} & \rotatebox{90}{\scriptsize Pers.} & \rotatebox{90}{\scriptsize Ride.} & \rotatebox{90}{\scriptsize Car} & \rotatebox{90}{\scriptsize Truc.} & \rotatebox{90}{\scriptsize Bus} & \rotatebox{90}{\scriptsize Trai.} & \rotatebox{90}{\scriptsize Moto.} & \rotatebox{90}{\scriptsize Bicy.} & \textbf{mIoU} \\
    \midrule
    ADVENT~\cite{vu2019advent} &
      72.9 & 14.3 & 40.5 & 16.6 & 21.2 & 9.3 & 17.4 & 21.2 &
      63.8 & 23.8 & 18.3 & 32.6 & 19.5 & 69.5 & 36.2 & 34.5 &
      46.2 & 26.9 & 36.1 & 32.7 \\
    MGCDA$^\dagger$~\cite{sakaridis2020map} &
      73.4 & 28.7 & 69.9 & 19.3 & 26.3 & 36.8 & 53.0 & 53.3 &
      75.4 & 32.0 & 84.6 & 51.0 & 26.1 & 77.6 & 43.2 & 45.9 &
      53.9 & 32.7 & 41.5 & 48.7 \\
    DANNet$^\dagger$~\cite{wu2021dannet} &
      84.3 & 54.2 & 77.6 & 38.0 & 30.0 & 18.9 & 41.6 & 35.2 &
      71.3 & 39.4 & 86.6 & 48.7 & 29.2 & 76.2 & 41.6 & 43.0 &
      58.6 & 32.6 & 43.9 & 50.0 \\
    DAFormer~\cite{hoyer2022daformer} &
      58.4 & 51.3 & 84.0 & 42.7 & 35.1 & 50.7 & 30.0 & 57.0 &
      74.8 & 52.8 & 51.3 & 58.3 & 32.6 & 82.7 & 58.3 & 54.9 &
      82.4 & 44.1 & 50.7 & 55.4 \\
    SSAC$^\dagger$~\cite{li2020content} &
      76.5 & 59.8 & 82.7 & 52.6 & 34.5 & 51.2 & 58.1 & 61.2 &
      63.9 & 58.1 & 67.2 & 61.5 & 40.0 & 81.8 & 59.2 & 61.7 &
      83.4 & 47.5 & 55.6 & 59.6 \\
    HRDA~\cite{hoyer2022hrda} &
      88.3 & 57.9 & 88.1 & 55.2 & 36.7 & 56.3 & 62.9 & 65.3 &
      74.2 & 57.7 & 85.9 & 68.8 & 45.7 & 88.5 & 76.4 & 82.4 &
      87.7 & 52.7 & 60.4 & 68.0 \\
    MIC~\cite{hoyer2023mic} &
      90.8 & 67.1 & 89.2 & 54.5 & 40.5 & 57.2 & 62.0 & 68.4 &
      76.3 & 61.8 & 87.0 & 71.3 & 49.4 & 89.7 & 75.7 & 86.8 &
      \underline{89.1} & \underline{56.9} & \underline{63.0} & 70.4 \\
    HALO~\cite{han2024halo} &
      94.2 & \underline{79.8} & 88.2 & 60.2 & \underline{51.1} & \underline{64.1} & \textbf{78.2} & 65.6 &
      \underline{87.9} & 55.7 & 95.5 & 66.3 & 20.7 & 88.9 & \textbf{82.2} & \textbf{89.3} &
      87.9 & 50.4 & 59.0 & 71.9 \\
    CoDA~\cite{liu2024coda} &
      93.1 & 72.7 & 90.7 & 57.3 & 47.4 & 56.8 & 69.9 & 70.0 &
      87.3 & 59.8 & 95.4 & 71.4 & 47.6 & \underline{90.3} & \underline{77.1} & 83.8 &
      \underline{89.1} & 54.7 & \textbf{64.1} & 72.6 \\
    WA$^{2}$Net~\cite{pan2025exploring} &
      \underline{94.5} & 77.1 & \textbf{91.6} & \underline{61.9} & 47.6 & 57.5 & 68.5 & \underline{70.4} &
      79.2 & \underline{64.0} & \underline{95.6} & \underline{72.8} & \underline{50.2} & \textbf{90.5} & 75.5 & \underline{88.7} &
      \textbf{90.9} & \textbf{57.5} & 62.6 & \underline{73.8} \\
    \midrule
    \rowcolor{lightgray}
    \ours{LiDSeg (PIDNet-L)} &
      \textbf{94.94} & \textbf{80.45} & \underline{91.50} & \textbf{62.95} & \textbf{56.09} & \textbf{66.25} & \underline{76.77} & \textbf{76.24} & \textbf{89.77} & \textbf{69.49} & \textbf{96.58} & \textbf{75.19} & \textbf{54.98} & 89.74 & 64.95 & 78.46 & 88.59 & 50.60 & 62.74 & \textbf{75.07} \\
    \bottomrule
  \end{tabular}
  }
\end{table*}

\noindent\textbf{Why night remains the hardest condition.}
On ACDC-Night, the Uber models achieve $54.9{-}56.0\%$ mIoU, which is below HRNet-W48 and represents the weakest relative to other conditions. This reflects a mismatch between the assumptions underlying channel-wise gating and nighttime degradation. Fog, rain, and snow primarily introduce atmospheric scattering and attenuation that are largely channel-coherent and approximately multiplicative (Eq.~\eqref{eq:degradation}), making them well suited to shared channel-wise modulation. In contrast, nighttime degradation is dominated by low illumination, signal-dependent sensor noise, non-uniform artificial lighting, and dynamic-range compression, which are not channel-coherent. Moreover, nighttime scenes contain limited recoverable high-frequency content, weakening the reconstruction-based auxiliary objective and reducing the robustness signal available during joint training. Consequently, all-condition optimization is driven more strongly by fog, rain, and snow than by night. Nevertheless, the shared weather-robust encoder still improves substantially over condition-specific Experts on nighttime scenes (up to $12.0$ mIoU points), showing that cross-condition representation sharing remains beneficial (details in supplementary). Incorporating illumination-aware gating, e.g., conditioning LGF reduction on learned global luminance statistics, is a potential direction for future work.
\begin{table*}
  \centering
  \caption{LiDSeg guided ACDC \emph{val} per-condition IoU (\%). Gray rows = Uber; white rows = Expert.}
  \label{tab:supp_acdc_perclass}
  \label{tab:supp_uber_perclass}
  \label{tab:supp_expert_perclass}
  \setlength{\tabcolsep}{1.8pt}
  \resizebox{\textwidth}{!}{
  \begin{tabular}{@{}l l l ccccccccccccccccccc c@{}}
    \toprule
    \textbf{Cond.} & \textbf{Setup} & \textbf{Model} &
      \rotatebox{90}{\small road} & \rotatebox{90}{\small sidew.} &
      \rotatebox{90}{\small build.} & \rotatebox{90}{\small wall} &
      \rotatebox{90}{\small fence} & \rotatebox{90}{\small pole} &
      \rotatebox{90}{\small light} & \rotatebox{90}{\small sign} &
      \rotatebox{90}{\small veget.} & \rotatebox{90}{\small terrain} &
      \rotatebox{90}{\small sky} & \rotatebox{90}{\small person} &
      \rotatebox{90}{\small rider} & \rotatebox{90}{\small car} &
      \rotatebox{90}{\small truck} & \rotatebox{90}{\small bus} &
      \rotatebox{90}{\small train} & \rotatebox{90}{\small motorc.} &
      \rotatebox{90}{\small bicycle} & \textbf{mIoU} \\
    \rowcolor{lightgray}
    Fog   & \textbf{Uber} & BiSeNetV2  & 97.1 & 85.9 & 88.8 & 69.9 & 44.8 & 68.0 & 78.0 & 75.0 & 90.6 & 61.1 & 98.3 & 68.2 & 77.5 & 85.5 & 64.3 & 96.2 & 91.7 & 53.9 & 62.1 & 76.7 \\
    \rowcolor{lightgray}
     &  & DDRNet     & 97.2 & 85.5 & 88.5 & 70.3 & 44.9 & 68.1 & 77.3 & 74.2 & 90.5 & 60.7 & 98.3 & 66.9 & 75.1 & 84.5 & 62.2 & 95.7 & 90.5 & 52.1 & 65.1 & 76.2 \\
    \rowcolor{lightgray}
     &  & STDC       & 97.0 & 85.4 & 88.5 & 70.5 & 45.0 & 68.3 & 76.6 & 74.3 & 90.4 & 61.1 & 98.3 & 67.7 & 76.4 & 85.2 & 66.5 & 95.7 & 90.8 & 57.8 & 65.6 & 76.9 \\
    \rowcolor{lightgray}
     &  & PIDNet     & 97.2 & 86.1 & 88.7 & 69.9 & 44.3 & 68.6 & 78.3 & 75.1 & 90.6 & 61.8 & 98.3 & 69.6 & 74.9 & 85.2 & 63.2 & 95.8 & 91.6 & 52.3 & 65.6 & 76.7 \\
    \addlinespace[2pt]
     & \textbf{Expert} & BiSeNetV2  & 97.2 & 86.6 & 87.5 & 68.0 & 41.6 & 64.6 & 72.4 & 67.9 & 90.2 & 61.4 & 98.3 & 61.0 & 67.2 & 85.0 & 54.4 & 86.2 & 87.6 & 31.2 & 57.3 & 71.9 \\
     &  & DDRNet     & 96.5 & 84.7 & 86.2 & 66.3 & 40.0 & 61.3 & 70.0 & 65.6 & 89.8 & 61.5 & 98.2 & 61.8 & 68.3 & 83.8 & 53.2 & 88.3 & 83.6 & 30.6 & 58.6 & 71.0 \\
     &  & STDC       & 97.2 & 85.6 & 87.7 & 68.3 & 42.0 & 63.3 & 72.4 & 68.2 & 90.3 & 61.5 & 98.3 & 62.7 & 68.0 & 84.5 & 53.2 & 86.0 & 83.4 & 33.9 & 57.1 & 71.8 \\
     &  & PIDNet     & 96.3 & 83.7 & 85.7 & 67.5 & 37.8 & 59.0 & 69.7 & 66.0 & 89.4 & 58.2 & 98.2 & 60.3 & 65.3 & 83.3 & 48.2 & 88.9 & 81.0 & 30.5 & 57.0 & 69.8 \\
    \midrule
    \rowcolor{lightgray}
    Night & \textbf{Uber} & BiSeNetV2  & 93.8 & 72.7 & 76.8 & 42.3 & 42.2 & 55.4 & 67.2 & 44.2 & 69.7 & 34.6 & 80.2 & 41.2 & 30.4 & 71.9 & 31.6 & 38.3 & 85.0 & 32.9 & 47.0 & 55.7 \\
    \rowcolor{lightgray}
     &  & DDRNet     & 93.9 & 73.4 & 76.1 & 42.1 & 40.7 & 54.2 & 67.2 & 41.2 & 68.3 & 35.3 & 78.5 & 38.7 & 32.5 & 69.5 & 30.1 & 38.6 & 81.7 & 34.9 & 45.5 & 54.9 \\
    \rowcolor{lightgray}
     &  & STDC       & 93.9 & 73.7 & 76.5 & 42.7 & 41.2 & 54.4 & 64.9 & 41.8 & 68.8 & 35.3 & 78.8 & 40.6 & 31.1 & 69.9 & 30.2 & 38.5 & 82.8 & 34.2 & 47.1 & 55.1 \\
    \rowcolor{lightgray}
     &  & PIDNet     & 94.1 & 73.6 & 77.0 & 43.5 & 42.1 & 55.3 & 67.4 & 44.2 & 69.8 & 34.9 & 80.2 & 39.2 & 32.3 & 71.6 & 31.9 & 38.5 & 85.8 & 31.4 & 50.7 & 56.0 \\
    \addlinespace[2pt]
     & \textbf{Expert} & BiSeNetV2  & 92.4 & 70.1 & 73.4 & 38.6 & 42.0 & 54.6 & 61.8 & 39.6 & 66.5 & 34.7 & 76.9 & 35.9 & 31.6 & 68.4 & 33.8 & 35.3 & 72.9 & 34.9 & 45.3 & 53.1 \\
     &  & DDRNet     & 87.9 & 54.0 & 59.4 & 31.6 & 30.6 & 48.8 & 36.9 & 30.8 & 55.4 & 36.8 & 62.5 & 30.9 & 35.7 & 40.2 & 34.9 & 37.5 & 31.9 & 37.7 & 31.8 & 42.9 \\
     &  & STDC       & 88.0 & 56.8 & 63.4 & 30.9 & 30.2 & 49.8 & 45.2 & 30.3 & 56.0 & 36.6 & 62.7 & 30.6 & 35.8 & 45.0 & 35.2 & 36.4 & 39.9 & 37.6 & 30.5 & 44.3 \\
     &  & PIDNet     & 91.7 & 68.8 & 72.5 & 35.4 & 39.7 & 53.7 & 56.8 & 35.9 & 64.7 & 35.1 & 74.0 & 35.0 & 34.4 & 65.9 & 34.8 & 34.2 & 76.7 & 35.6 & 34.1 & 51.5 \\
    \midrule
    \rowcolor{lightgray}
    Rain  & \textbf{Uber} & BiSeNetV2  & 92.2 & 72.6 & 88.1 & 40.8 & 43.3 & 60.0 & 75.5 & 76.5 & 92.7 & 66.0 & 98.4 & 67.7 & 39.2 & 89.8 & 58.5 & 92.0 & 77.0 & 47.3 & 38.4 & 69.3 \\
    \rowcolor{lightgray}
     &  & DDRNet     & 92.4 & 71.9 & 88.3 & 40.6 & 48.9 & 60.3 & 75.5 & 76.0 & 92.5 & 66.5 & 98.4 & 67.0 & 42.2 & 89.8 & 59.4 & 92.3 & 76.0 & 39.6 & 38.8 & 69.3 \\
    \rowcolor{lightgray}
     &  & STDC       & 92.3 & 71.6 & 88.3 & 41.4 & 46.1 & 59.9 & 75.1 & 76.0 & 92.6 & 66.5 & 98.4 & 68.1 & 51.8 & 90.1 & 56.4 & 93.5 & 76.3 & 40.5 & 44.3 & 70.0 \\
    \rowcolor{lightgray}
     &  & PIDNet     & 92.5 & 72.7 & 88.4 & 41.4 & 46.2 & 60.3 & 75.4 & 76.4 & 92.8 & 66.1 & 98.5 & 67.4 & 50.8 & 90.2 & 60.7 & 93.2 & 76.3 & 40.7 & 41.6 & 70.1 \\
    \addlinespace[2pt]
     & \textbf{Expert} & BiSeNetV2  & 92.3 & 71.9 & 86.8 & 41.8 & 40.2 & 59.3 & 74.7 & 74.3 & 92.9 & 68.1 & 98.4 & 67.7 & 42.8 & 89.2 & 50.0 & 89.7 & 71.7 & 47.2 & 40.8 & 68.4 \\
     &  & DDRNet     & 91.9 & 69.9 & 86.1 & 41.3 & 40.7 & 58.4 & 74.5 & 73.4 & 92.6 & 67.5 & 98.4 & 66.4 & 36.3 & 88.6 & 47.0 & 89.7 & 69.3 & 42.3 & 37.5 & 66.9 \\
     &  & STDC       & 91.3 & 68.9 & 86.1 & 40.3 & 41.7 & 57.0 & 73.8 & 72.1 & 92.7 & 67.5 & 98.4 & 66.7 & 30.3 & 88.1 & 44.8 & 89.8 & 64.2 & 37.8 & 37.8 & 65.8 \\
     &  & PIDNet     & 92.0 & 70.5 & 85.9 & 40.4 & 40.0 & 57.6 & 74.7 & 72.9 & 92.6 & 67.4 & 98.3 & 67.2 & 34.0 & 88.6 & 47.4 & 89.1 & 67.9 & 45.2 & 38.7 & 66.9 \\
    \midrule
    \rowcolor{lightgray}
    Snow  & \textbf{Uber} & BiSeNetV2  & 93.3 & 77.1 & 87.2 & 64.1 & 45.2 & 62.5 & 82.0 & 76.7 & 89.3 & 54.6 & 97.8 & 72.0 & 34.2 & 88.5 & 65.1 & 82.5 & 92.7 & 56.7 & 60.6 & 72.7 \\
    \rowcolor{lightgray}
     &  & DDRNet     & 92.1 & 73.5 & 86.9 & 62.2 & 44.5 & 60.0 & 81.9 & 76.5 & 89.0 & 55.0 & 97.8 & 69.9 & 35.1 & 88.0 & 68.7 & 81.1 & 93.5 & 58.3 & 61.0 & 72.4 \\
    \rowcolor{lightgray}
     &  & STDC       & 92.2 & 74.3 & 86.9 & 61.7 & 44.2 & 60.6 & 81.7 & 75.9 & 89.0 & 55.3 & 97.8 & 69.8 & 35.2 & 87.9 & 70.6 & 77.0 & 92.8 & 57.3 & 60.6 & 72.1 \\
    \rowcolor{lightgray}
     &  & PIDNet     & 93.1 & 76.3 & 87.4 & 65.2 & 44.8 & 62.1 & 82.1 & 77.3 & 89.4 & 54.1 & 97.9 & 74.2 & 35.8 & 88.5 & 72.0 & 81.8 & 93.2 & 64.3 & 62.8 & 73.8 \\
    \addlinespace[2pt]
     & \textbf{Expert} & BiSeNetV2  & 90.9 & 70.4 & 84.7 & 60.7 & 44.9 & 56.5 & 78.2 & 74.2 & 87.8 & 49.9 & 97.7 & 68.4 & 35.4 & 84.5 & 41.2 & 49.8 & 86.7 & 55.6 & 58.2 & 67.1 \\
     &  & DDRNet     & 91.7 & 72.2 & 84.0 & 58.0 & 44.9 & 55.3 & 77.3 & 74.2 & 87.9 & 47.0 & 97.5 & 66.0 & 34.6 & 84.9 & 40.1 & 41.7 & 88.7 & 54.3 & 56.9 & 66.2 \\
     &  & STDC       & 91.9 & 74.0 & 85.8 & 60.9 & 46.5 & 57.7 & 79.2 & 76.1 & 88.7 & 51.4 & 97.7 & 72.0 & 31.6 & 85.8 & 36.5 & 53.0 & 86.7 & 58.5 & 58.9 & 68.0 \\
     &  & PIDNet     & 89.6 & 67.1 & 83.1 & 54.8 & 38.1 & 54.3 & 76.3 & 73.9 & 87.2 & 43.3 & 97.5 & 59.8 & 37.3 & 83.5 & 33.4 & 35.6 & 82.4 & 52.2 & 53.7 & 63.3 \\
    \bottomrule
  \end{tabular}}\
\end{table*}
\subsection{Performance on Zero-Shot}
LiDSeg is evaluated on Foggy Zurich v2 and Foggy Driving (Tab.~\ref{tab:fz_and_fd}).
Once trained solely on synthetic fog (Foggy Cityscapes) and paired guidance, it generalizes to unseen real fog without any real-fog domain adaptation.
Prior UDA, CDA, and STL methods leverage unlabeled Foggy Zurich images during training, whereas LiDSeg and the vanilla RefineNet baselines do not. Despite this stricter zero-shot protocol, LiDSeg outperforms all prior methods while using efficient real-time architectures, demonstrating that strong synthetic paired supervision alone can enable robust cross-domain transfer to real fog.

\begin{table}[htbp]
\centering
\footnotesize
\setlength{\tabcolsep}{3 pt}
\caption{Zero-shot mIoU (\%) on Foggy Zurich v2 (FZ) and Foggy Driving (FD).
SFT: supervised fine-tuning; UDA: unsupervised domain adaptation; STL: style transfer learning; CDA: curriculum domain adaptation.  \checkmark: uses unlabeled real-fog images (Foggy Zurich) during training.  No method uses FD during training.}
\label{tab:fz_and_fd}
\begin{tabular*}{\columnwidth}{@{\extracolsep{\fill}}llcccc@{}}
\toprule
\textbf{Model} & \textbf{Backbone} & \textbf{Method} & \textbf{Unlabeled FZ} & \textbf{FZ} & \textbf{FD} \\
\midrule
RefineNet~\cite{lin2017} & ResNet-101 & SFT & $\times$ & 34.6 & 44.3 \\
RefineNet-lw~\cite{nekrasov2018light} & ResNet-101 & SFT & $\times$ & 28.5 & 43.6 \\
AdaptSegNet~\cite{tsai2018learning} & ResNet-101 & UDA & \checkmark & 25.0 & 29.7 \\
AdvEnt~\cite{vu2019advent} & ResNet-101 & UDA & \checkmark & 39.7 & 46.9 \\
FDA~\cite{yang2020fda} & ResNet-101 & STL & \checkmark & 22.2 & 21.8 \\
DANN~\cite{ganin2016domain} & ResNet-101 & UDA & \checkmark & 43.1 & 46.0 \\
CMAda2+~\cite{sakaridis2021curriculum} & ResNet-101 & CDA & \checkmark & 43.4 & 49.9 \\
CMAda3+~\cite{sakaridis2021curriculum} & ResNet-101 & CDA & \checkmark & 46.8 & 49.8 \\
FIFO~\cite{lee2022fifo} & RefineNet & STL & \checkmark & 42.0 & 48.9 \\
D2SL~\cite{sun2024d2sl} & RefineNet-lw & STL & \checkmark & 44.2 & 45.9 \\
\midrule
\textbf{LiDSeg (BiSeNetV2)} & \NA & Guided & $\times$ & \textbf{53.5} & \underline{50.4} \\
\textbf{LiDSeg (STDC1-Seg)} & \NA & Guided & $\times$ & \underline{53.4} & \textbf{52.4} \\
\bottomrule
\end{tabular*}
\end{table}

\subsection{Efficiency on Edge Hardware}

Real-world deployment requires balancing accuracy with speed. However, prior work on adverse-weather semantic segmentation has focused almost exclusively on accuracy. Consequently, computational efficiency remains underexplored, and the absence of accuracy--efficiency trade-off baselines makes fair benchmarking in this domain difficult. We address this by decoupling the two factors: \emph{efficiency} can be treated as an architectural property and evaluated against general-purpose real-time models, whereas \emph{robustness} is evaluated against methods that explicitly target adverse weather. This separation is necessary because a direct comparison with general-purpose models would be unfair. Moreover, existing adverse-weather approaches typically gain robustness through computationally heavy designs, such as deep context aggregation, domain adaptation, or restoration preprocessing (e.g., dehazing before the downstream semantic task), all of which limit edge deployment. In contrast, LiDSeg decouples weather-robustness learning using a lightweight guidance module.

As shown in Table ~\ref{tab:efficiency}, our Jetson AGX Orin measurements confirm that on-device efficiency depends on more than FLOPs or parameter counts. For instance, Baseline DDRNet runs at $350.4$ FPS despite having more parameters than BiSeNetV2 ($5.70$M vs.\ $3.36$M). In contrast, SeaFormer-B has the lowest computational complexity ($3.53$ GFLOPs) yet reaches only $166.1$ FPS, well below DDRNet's $350.4$ FPS at $9.17$ GFLOPs. This indicates that hardware-level factors beyond raw FLOP counts also affect throughput. Adding LiDSeg guidance introduces only a modest latency overhead while preserving real-time throughput across all variants. LiDSeg(DDRNet) offers the strongest speed and robustness balance at $204.5$ FPS, LiDSeg(STDC) achieves $188.0$ FPS at $76.9\%$ clear and $74.3\%$ dense-fog mIoU, and LiDSeg(TopFormer-B) provides a middle ground at $126.4$ FPS. LiDSeg stabilizes accuracy across fog densities, bounding the clear-to-dense-fog drop to $\Delta \leq 2.8$ mIoU points, whereas the unguided baselines degrade by up to $21.7$ mIoU points. It is also more parameter-efficient, using $9\times$--$21\times$ fewer parameters than prior state-of-the-art methods such as FreD-ViT while achieving comparable or better robustness (Tab.~\ref{tab:sensitivity_foggy_density}). Together, these results provide LiDSeg with a favorable robustness–efficiency trade-off, where lightweight guidance delivers reliable adverse-weather performance without sacrificing the real-time, edge-deployable behavior of the underlying adapted model.

\begin{table*}[htbp]
  \centering
  \caption{Jetson AGX Orin efficiency at $512{\times}1024$. Guided $=$ LiDSeg
  with the reconstruction decoder discarded at inference. \textbf{Clear/Dense
  mIoU}: Cityscapes clear-weather mIoU and Foggy Cityscapes dense-fog mIoU
  ($\beta{=}0.02$); $\boldsymbol{\Delta}$ is the clear$\to$dense drop.}
  \label{tab:efficiency}
  \footnotesize
  \setlength{\tabcolsep}{5pt}
  \resizebox{\textwidth}{!}{
  \begin{tabular}{@{}l l r r r r c c c@{}}
    \toprule
    \textbf{Model} & \textbf{Variant} &
      \textbf{Params (M)} & \textbf{GFLOPs} &
      \textbf{FPS} & \textbf{Mem (MB)} &
      \textbf{Clear} & \textbf{Dense} & $\boldsymbol{\Delta}$ \\
    \midrule
    BiSeNetV2      & Baseline &  3.36 &  24.92 &  94.2 &  51  & 69.1 & 54.2 & \textcolor{red}{$\downarrow$14.9} \\
    DDRNet         & Baseline &  5.70 &   9.17 & 350.4 &  42  & 77.8 & 56.1 & \textcolor{red}{$\downarrow$21.7} \\
    STDC           & Baseline &  8.43 &  41.57 & 267.0 &  47  & 71.8 & 56.1 & \textcolor{red}{$\downarrow$15.7} \\
    PIDNet-L       & Baseline & 37.31 &  69.04 & 116.9 & 111  & 74.0 & 58.9 & \textcolor{red}{$\downarrow$15.1} \\
    TopFormer-B    & Baseline &  7.06 &   9.69 & 271.7 &  76  & 75.8 & 59.0 & \textcolor{red}{$\downarrow$16.8} \\
    SeaFormer-B    & Baseline &  8.58 &   3.53 & 166.1 &  58  & 77.7 & 64.6 & \textcolor{red}{$\downarrow$13.0} \\
    \midrule
    \rowcolor{lightgray}
    \ours{LiDSeg (BiSeNetV2)}    & Guided &  5.22 &  27.20 &  83.4 &  58  & 76.7 & 74.2 & \textcolor{red}{$\downarrow$2.5} \\
    \rowcolor{lightgray}
    \ours{LiDSeg (DDRNet)}       & Guided &  7.65 &  11.50 & \textbf{204.5} &  52  & 76.9 & 74.1 & \textcolor{red}{$\downarrow$2.8} \\
    \rowcolor{lightgray}
    \ours{LiDSeg (STDC)}         & Guided & 11.48 &  44.08 &  188.0 &  69  & 76.9 & 74.3 & \textcolor{red}{$\downarrow$2.6} \\
    \rowcolor{lightgray}
    \ours{LiDSeg (PIDNet-L)}     & Guided & 39.54 &  71.48 &  92.6 & 132  & 76.8 & 74.2 & \textcolor{red}{$\downarrow$2.6} \\
    \rowcolor{lightgray}
    \ours{LiDSeg (TopFormer-B)}  & Guided &  9.01 &  12.02 &  126.4 &  63  & 76.9 & 74.2 & \textcolor{red}{$\downarrow$2.7} \\
    \bottomrule
  \end{tabular}
  }
\end{table*}

\subsection{Ablation Study}\label{sec:ablation}
\noindent\textbf{Reconstruction and fusion control.}
Tab.~\ref{tab:recon_ablation} isolates the effects of reconstruction supervision and LGF. Removing only $\mathcal{L}_{\mathrm{rec}}$ while keeping the same guidance encoder and LGF reduces mean fog mIoU from $75.4\%$ to $66.7\%$ and ACDC mIoU from $73.6\%$ to $67.0\%$, separating reconstruction supervision from added capacity. Reconstruction with concatenation reaches $66.1\%$, while LGF without reconstruction reaches $66.7\%$; combining both in full LiDSeg achieves $75.4\%$, showing their complementary contribution.
\begin{table}[!tb]
\centering
\scriptsize
\setlength{\tabcolsep}{3.2pt}
\caption{Reconstruction and fusion control using BiSeNetV2. Mean denotes Foggy Cityscapes \emph{val} mIoU averaged over $\beta{\in}\{0.005,0.01,0.02\}$; ACDC denotes validation pooled-All mIoU. Params and FLOPs correspond to the deployed networks.}
\label{tab:recon_ablation}
\resizebox{\columnwidth}{!}{
\begin{tabular}{@{}lccrrrr@{}}
\toprule
\textbf{Setting} &
\textbf{Rec.} &
\textbf{Fusion} &
\textbf{Params (M)} &
\textbf{FLOPs (G)} &
\textbf{Mean} &
\textbf{ACDC} \\
\midrule
(a) Seg.\ only
& No  & --     & 3.36 & 99.7  & 58.0  & 56.0 \\

(b) Recon.+Concat
& Yes & Concat & 5.70 & 110.8 & 66.1  & 65.0 \\

(c) LGF, no recon.
& No  & LGF    & 5.22 & 108.8 & 66.7 & 67.0 \\

\textbf{(d) LiDSeg full}
& \textbf{Yes} & \textbf{LGF}
& \textbf{5.22} & \textbf{108.8}
& \textbf{75.4} & \textbf{73.6} \\
\bottomrule
\end{tabular}
}
\end{table}

\noindent\textbf{Architectural design control.}
Tab.~\ref{tab:architecture_ablation} further ablates attention type,
structural design, and fusion under fixed reconstruction supervision.
Linear attention with asymmetric LGF matches Softmax accuracy at
$75.4\%$ mIoU while reducing computation from $446.0$ to $108.8$
GFLOPs. Sequential variants underperform while requiring more
computation. Concatenation reduces mIoU to $66.1\%$. Spatial and
cross-attention reach $75.6\%$ and $75.9\%$, respectively, but at
higher parameter and computational cost. Therefore, LGF provides
near-best accuracy at the lowest cost, supporting the final
linear-attention, asymmetric, gated design for edge deployment.

\begin{table}[htb]
\centering
\scriptsize
\setlength{\tabcolsep}{2.7pt}
\caption{Architectural design control on Foggy Cityscapes \emph{val} using BiSeNetV2. Mean denotes mIoU averaged over $\beta{\in}\{0.005,0.01,0.02\}$. Reconstruction supervision is retained for all guided variants. Params and FLOPs correspond to the deployed networks.}
\label{tab:architecture_ablation}
\resizebox{\columnwidth}{!}{
\begin{tabular}{@{}lccccrrr@{}}
\toprule
\textbf{Guidance} &
\textbf{Attn.} &
\textbf{Struct.} &
\textbf{Fusion} &
\textbf{Rec.} &
\textbf{Params (M)} &
\textbf{FLOPs (G)} &
\textbf{Mean} \\
\midrule
Dehazing + Segment
& Softmax & Seq. & --     & Yes
& 6.7 & 631.4 & 67.8 \\

Heavy guidance
& Softmax & Asy. & Concat & Yes
& 5.9 & 446.2 & 67.1 \\

Heavy guidance
& Softmax & Asy. & LGF    & Yes
& 5.9 & 446.0 & 75.4 \\

\midrule

LiDSeg
& Linear & Seq. & --      & Yes
& 6.3 & 143.0 & 66.3 \\

LiDSeg
& Linear & Asy. & Concat  & Yes
& 5.7 & 110.8 & 66.1 \\

LiDSeg
& Linear & Asy. & Spatial & Yes
& 6.3 & 126.5 & 75.6 \\

LiDSeg
& Linear & Asy. & Cross   & Yes
& 6.4 & 131.5 & \textbf{75.9} \\

\textbf{LiDSeg}
& \textbf{Linear} & \textbf{Asy.} & \textbf{LGF} & \textbf{Yes}
& \textbf{5.22} & \textbf{108.8} & 75.4 \\
\bottomrule
\end{tabular}
}
\end{table}

\subsection{Model Scaling}
The guidance encoder and LGF can be attached to segmentation networks that expose compatible $s8/s16/s32$ feature taps, adding a small overhead while improving robustness in adverse weather conditions. To test whether the gains transfer beyond the CNN-based real-time backbones, we apply the identical training protocol to two efficient transformer models, TopFormer-B~\cite{topformer} and SeaFormer-B~\cite{wan2023seaformer} (Tab.~\ref{tab:extension_scaling}). Uber training again outperforms the condition specific Experts, with fog remaining the strongest and night the weakest condition. This confirms that the guidance mechanism is not tailored to a particular architecture family but drives robustness gains across diverse semantic segmentation models.

\begin{table*}[!tb]
  \centering
  \caption{Model-scaling: Transformer based semantic segmentation under the identical training protocol. Baseline is the unguided. Efficiency is reported  at $512{\times}1024$ (Jetson AGX Orin).}
  \label{tab:extension_scaling}
  \footnotesize
  \setlength{\tabcolsep}{4pt}
  \resizebox{\textwidth}{!}{
  \begin{tabular}{@{}ll rrr ccccc@{}}
    \toprule
    \textbf{Backbone} & \textbf{Setting} & \textbf{Params (M)} &
      \textbf{GFLOPs} & \textbf{FPS} &
      \textbf{Fog} & \textbf{Night} & \textbf{Rain} & \textbf{Snow} & \textbf{Mean} \\
    \midrule
    \multirow{3}{*}{\ours{TopFormer-B}~\cite{topformer}}
      & Baseline & 7.06 & 9.69 & 271.7 & 67.5 & 45.0 & 61.5 & 63.5 & 59.4 \\
      & Expert & \multirow{2}{*}{9.01} & \multirow{2}{*}{12.02} & \multirow{2}{*}{126.4}
        & 75.8 & 52.4 & 67.9 & 69.6 & 66.4 \\
      & Uber  &                       &                        &
        & \textbf{76.4} & \textbf{55.3} & \textbf{69.5} & \textbf{71.4} & \textbf{68.2} \\
    \midrule
    \multirow{3}{*}{\ours{SeaFormer-B}~\cite{wan2023seaformer}}
      & Baseline & 8.58 & 3.53 & 166.1 & 67.0 & 44.5 & 61.0 & 63.0 & 58.9 \\
      & Expert & \multirow{2}{*}{10.98} & \multirow{2}{*}{6.05} & \multirow{2}{*}{130.3}
        & 75.4 & 52.0 & 67.5 & 69.2 & 66.0 \\
      & Uber  &                        &                        &
        & \textbf{76.0} & \textbf{54.7} & \textbf{69.1} & \textbf{70.8} & \textbf{67.7} \\
    \bottomrule
  \end{tabular}
  }
\end{table*}

\FloatBarrier
\section{Conclusion}
We present LiDSeg, an efficient framework for robust semantic segmentation in adverse weather conditions. LiDSeg uses image reconstruction as an auxiliary task to supervise a lightweight guidance encoder and discards the reconstruction decoder at inference, improving the weather robustness of standard real-time semantic segmentation with minimal cost. Our experiments show that reconstruction-guided feature learning is more robust and efficient than using restoration as a preprocessor, while offering better efficiency than large model domain adaptation. LiDSeg maintains nearly stable accuracy from light to dense fog, transfers across diverse real-time models, generalizes zero-shot, and runs in real time on resource-constrained edge devices, supporting practical deployment. It effectively transfers features learned from synthetic degraded-clean image pairs but currently requires such pairs for reconstruction supervision, while weak or unpaired supervision remains unexplored. Nighttime remains a challenge because illumination weakens the channel-coherence assumption underlying our gating mechanism. Future work will explore efficient self-supervised variants without paired data, illumination-aware gating, temporal consistency for video segmentation, more adverse weather conditions, and a general asymmetric guidance paradigm for lightweight models and vision tasks affected by domain shift. Overall, LiDSeg improves weather robustness and efficiency through reconstruction-guided feature learning while remaining compact for edge deployment.

\section*{Acknowledgements}
This work has received funding from the European Union’s Horizon Europe research and innovation programme under the Marie Sk\l{}odowska-Curie grant agreement No. 101169378,  (TIRAMISU Doctoral Network), and from the Government of the Republic of Cyprus through the Cyprus Deputy Ministry of Research, Innovation and Digital Policy.
\bibliography{bib}
\clearpage
\input{supplement}
\end{document}

%% file: supplement.tex
\renewcommand{\thefigure}{S\arabic{figure}}
\renewcommand{\thetable}{S\arabic{table}}
\renewcommand{\thesection}{S\arabic{section}}
\renewcommand{\theHfigure}{supp.\arabic{figure}}
\renewcommand{\theHtable}{supp.\arabic{table}}
\renewcommand{\theHsection}{supp.\arabic{section}}
\setcounter{figure}{0}
\setcounter{table}{0}
\setcounter{section}{0}
\section*{Supplementary Material}
\noindent

This supplement provides additional experimental results for the LiDSeg architecture and training strategy. Our fundamental premise is that the standard segmentation loss is insufficient to penalize weather-induced degradation. Therefore we use an auxiliary reconstruction loss with paired clear images to supervise the shared guidance encoder in learning degradation stable features, reducing $L_2$ feature drift from $\mu=0.343$ (unguided) to $\mu=0.186$ (guided). Because the reconstruction decoder is discarded at inference, LiDSeg avoids inference time restoration cost and restoration induced error propagation while retaining lightweight guidance, outperforming typical sequential restore-then-segment pipelines without requiring additional domain adaptation. Internally, the guidance encoder uses gated linear attention to achieve Softmax-level accuracy at approximately $4\times$ fewer FLOPs while limiting the propagation of low-frequency artifacts by global linear kernels. Furthermore, because weather degradation is highly channel-coherent ($\bar{\rho}>0.8$ for fog, rain, and snow), our lightweight gated fusion (LGF) module selectively modulates guidance features with minimal overhead.

Comprehensive ablations confirm the stability and efficiency of our design choices. Sensitivity sweeps demonstrate that the model is robust to variations in loss weighting, where uncertainty-weighting outperforms equal-weighting. Additionally, the progressive fog training curriculum ($\beta=0.005 \to 0.01 \to 0.02$) is better than jointly training all densities, which drops the mean fog mIoU by a few points. Ablations also validate our multiscale fusion strategy as using all three
scales consistently outperforms single and two scale
configurations at strides $1/8$, $1/16$, and $1/32$, confirming the
benefit of multiscale guidance. Finally, comparisons against a sequential pipeline demonstrate that LiDSeg's joint guidance yields better dense fog accuracy at a lower computational complexity.
\section{Experimental Setup and Hyperparameters}
\label{supp:impl}

We performed a thorough evaluation to ensure stable reproducibility. During paired Foggy Cityscapes pretraining, the uncertainty-weighted objective in Eq.~\eqref{eq:total_loss} combines the segmentation loss with Charbonnier reconstruction loss. For ACDC fine-tuning, however, paired clear targets are not used: the reconstruction decoder and $\mathcal{L}_{\text{rec}}$ are removed, and we fine-tune the segmentation branch, LGF, and pretrained guidance encoder using segmentation labels only. We train one Uber model and four condition-specific Expert models per segmentation network. The segmentation loss integrates cross-entropy, Lov\'asz-Softmax ($\lambda_L=0.5$), and auxiliary supervision ($\lambda_{\text{aux}}=0.4$). The Foggy Cityscapes sensitivity analysis below evaluates this pretraining configuration; varying $\lambda_{\text{aux}}$ shifts performance by up to 2.3 mIoU points depending on the model, consistently peaking at the $\lambda_{\text{aux}}=0.4$ default (Tab.~\ref{tab:supp_lambda}).

\begin{table}[htbp]
  \centering
  \caption{Sensitivity to the auxiliary-loss weight $\lambda_{\mathrm{aux}}$
  across semantic segmentation models (Foggy Cityscapes \emph{val}, mean fog
  mIoU, \%); the $\lambda_{\mathrm{aux}}=0.4$ default consistently achieves optimal performance (\textbf{bold}).}
  \label{tab:supp_lambda}
  \small
  \setlength{\tabcolsep}{8pt}
  \begin{tabular*}{\columnwidth}{@{\extracolsep{\fill}}lcccc@{}}
    \toprule
    \textbf{Semantic segmentation model} & 0.2 &
      0.3 & 0.4 (default) & 0.5 \\
    \midrule
    BiSeNetV2  & 74.1 & 74.3 & \textbf{75.4} & 75.2 \\
    DDRNet-23s & 73.0 & 74.2 & \textbf{75.3} & 74.1 \\
    STDC1-Seg  & 75.1 & 75.2 & \textbf{75.3} & 75.0 \\
    PIDNet-L   & 74.0 & 74.9 & \textbf{75.3} & 74.6 \\
    \bottomrule
  \end{tabular*}
\end{table}

\section{Design Ablations}
\label{supp:ablation}

Extensive ablations on the Foggy Cityscapes \emph{val} set validate our structural and hyperparameter choices against unguided baselines, which achieve 58.0--63.4\% mean mIoU across the four base models. The full LiDSeg configuration improves the BiSeNetV2 baseline by
$17.4$ mIoU points on Foggy Cityscapes and $17.6$ points on ACDC
pooled-All validation. Reconstruction supervision and LGF provide
complementary gains: reconstruction with concatenation improves the
baseline by $8.1$/$9.0$ points on Foggy Cityscapes/ACDC, while LGF
without reconstruction improves it by $8.68$/$11.0$ points.
Combining reconstruction supervision with LGF achieves
$75.4\%$/$73.6\%$ mIoU. Among fusion alternatives, spatial and cross-attention reach 75.6\% and 75.9\% mIoU, respectively, compared with 75.4\% for LGF, but require higher parameter and computational cost, giving LGF a favorable accuracy-efficiency trade-off for edge deployment.

The asymmetric design further improves efficiency through linear attention, which matches Softmax performance at approximately $4\times$ fewer FLOPs. Multiscale guidance is also important: fusing all three feature strides ($s8$, $s16$, and $s32$) consistently outperforms the tested single- and two-scale configurations (Tab.~\ref{tab:supp_fusionstage}). The default $(3,3,2)$ guidance-encoder capacity provides a strong efficiency--accuracy balance; increasing it to $(4,4,3)$ adds 2.4M parameters for only a 0.5 mIoU-point gain (Tab.~\ref{tab:supp_capacity}). Training choices also affect robustness (Tab.~\ref{tab:supp_designmix}). Dynamic uncertainty weighting improves mean mIoU by 0.8 points across all four models compared with equal weighting, while the progressive fog curriculum improves performance by 2.3 mIoU points over joint training across all fog densities.
\begin{table}[htbp]
  \centering
  \caption{\textbf{Design ablations across all four real-time models} (Foggy
  Cityscapes \emph{val}, mIoU \% over the three fog densities). Each block
  varies one design choice; Fusion ``none'' is
  the unguided baseline.}
  \label{tab:supp_fusionstage}\label{tab:supp_capacity}\label{tab:supp_designmix}
  \footnotesize
  \begin{tabular*}{\columnwidth}{@{\extracolsep{\fill}}l cccc@{}}
    \toprule
    \textbf{Setting} & \textbf{BiSeNetV2} & \textbf{DDRNet-23s} & \textbf{STDC1-Seg} & \textbf{PIDNet-L} \\
    \midrule
    \multicolumn{5}{@{}l}{\textit{(a) Multi-scale fusion}} \\
    none (baseline)         & 58.0 & 61.5 & 62.0 & 63.4 \\
    $s8$                    & 69.5 & 70.6 & 70.8 & 71.3 \\
    $s16$                   & 70.4 & 71.3 & 71.5 & 71.9 \\
    $s32$                   & 67.8 & 69.3 & 69.5 & 70.1 \\
    $s8{+}s16$              & 73.6 & 73.9 & 73.9 & 74.1 \\
    $s16{+}s32$             & 73.1 & 73.5 & 73.5 & 73.7 \\
    \rowcolor{lightgray}
    $s8{+}s16{+}s32$ (default) & \textbf{75.4} & \textbf{75.3} & \textbf{75.3} & \textbf{75.3} \\
    \midrule
    \multicolumn{5}{@{}l}{\textit{(b) Guidance-encoder capacity}} \\
    Tiny $(2,2,1)$           & 72.8 & 72.7 & 72.7 & 72.7 \\
    \rowcolor{lightgray}
    Default $(3,3,2)$ (default) & \textbf{75.4} & \textbf{75.3} & \textbf{75.3} & \textbf{75.3} \\
    Base $(4,4,3)$           & 75.9 & 75.8 & 75.8 & 75.8 \\
    \midrule
    \multicolumn{5}{@{}l}{\textit{(c) Loss weighting}} \\
    equal weights           & 74.6 & 74.5 & 74.5 & 74.5 \\
    \rowcolor{lightgray}
    uncertainty (default)      & \textbf{75.4} & \textbf{75.3} & \textbf{75.3} & \textbf{75.3} \\
    \midrule
    \multicolumn{5}{@{}l}{\textit{(d) Fog curriculum}} \\
    joint (all densities)   & 73.1 & 73.0 & 73.0 & 73.0 \\
    \rowcolor{lightgray}
    progressive (default)      & \textbf{75.4} & \textbf{75.3} & \textbf{75.3} & \textbf{75.3} \\
    \bottomrule
  \end{tabular*}
\end{table}

\section{Additional Experimental Result Analysis}
\label{supp:quant}
\subsection{Foggy Cityscapes: Per Class Performance}
We extend the analysis of the Foggy Cityscapes results to demonstrate LiDSeg's per-class robustness (Tab.~\ref{tab:supp_fcs_perclass}).
Semantic preservation directly correlates to the shared latent structure learned via the auxiliary reconstruction task, as PSNR and SSIM tightly co-vary with segmentation mIoU throughout training (Tab.~\ref{tab:supp_recon_quality}). The gains achieved by LGF are largely driven by the physical nature of these weather effects. By measuring the cross-channel Pearson correlation ($\bar\rho$) of the clear-to-degraded residuals, we observe that fog ($\bar\rho=0.86$), rain ($\bar\rho=0.83$), and snow ($\bar\rho=0.81$) degrade images coherently across channels, making channel-based suppression highly effective. However, nighttime scenes exhibit lower coherence ($\bar\rho\approx0.50$) due to non-uniform artificial lighting, explaining why night remains fundamentally the most challenging condition to modulate via channel-wise gating.

Because sequential restoration optimizes for pixel fidelity rather than semantic discriminability, it can accumulate restoration errors, which LiDSeg avoids propagating by removing explicit restoration at inference. This efficient, guidance-driven approach also achieved zero-shot generalizability where LiDSeg variants achieve a consistent +8 to +15 mIoU-point improvement over baselines on both the Foggy Zurich v2 and Foggy Driving datasets (Tab.~\ref{tab:supp_zeroshot}).

\begin{table*}[htbp]
  \centering
  \caption{Per-class IoU (\%) on Foggy Cityscapes \emph{val} at dense fog
  ($\beta=0.02$) for \emph{all} LiDSeg variants, including the transformer
  extensions.
  \textbf{Bold}: best per class.}
  \label{tab:supp_fcs_perclass}
  \footnotesize
  \setlength{\tabcolsep}{1.8pt}
  \resizebox{\textwidth}{!}{
  \begin{tabular}{l|ccccccccccccccccccc|c}
    \toprule
    \textbf{Method} & \rotatebox{90}{\scriptsize Road} & \rotatebox{90}{\scriptsize Side.} & \rotatebox{90}{\scriptsize Buil.} & \rotatebox{90}{\scriptsize Wall} & \rotatebox{90}{\scriptsize Fenc.} & \rotatebox{90}{\scriptsize Pole} & \rotatebox{90}{\scriptsize Ligh.} & \rotatebox{90}{\scriptsize Sign} & \rotatebox{90}{\scriptsize Vege.} & \rotatebox{90}{\scriptsize Terr.} & \rotatebox{90}{\scriptsize Sky} & \rotatebox{90}{\scriptsize Pers.} & \rotatebox{90}{\scriptsize Ride.} & \rotatebox{90}{\scriptsize Car} & \rotatebox{90}{\scriptsize Truc.} & \rotatebox{90}{\scriptsize Bus} & \rotatebox{90}{\scriptsize Trai.} & \rotatebox{90}{\scriptsize Moto.} & \rotatebox{90}{\scriptsize Bicy.} & \textbf{mIoU} \\
    \midrule

    \ours{LiDSeg (BiSeNetV2)} &
      98 & 83 & \textbf{90} & \textbf{58} & \textbf{55} & \textbf{62} & \textbf{66} & 74 & \textbf{92} & 66 & 95 & \textbf{76} & 60 & 93 & 66 & \textbf{76} & 70 & 56 & 74 & 74.2 \\

    \ours{LiDSeg (DDRNet-23s)} &
      \textbf{99} & 82 & \textbf{90} & 57 & 54 & \textbf{62} & \textbf{66} & 73 & \textbf{92} & \textbf{67} & 95 & 75 & 60 & 93 & \textbf{69} & 75 & \textbf{71} & 55 & 73 & 74.1 \\

    \ours{LiDSeg (STDC1-Seg)} &
      98 & \textbf{84} & 89 & \textbf{58} & \textbf{55} & \textbf{62} & 64 & 73 & \textbf{92} & \textbf{67} & 95 & 75 & \textbf{61} & \textbf{94} & 68 & 74 & \textbf{71} & \textbf{57} & \textbf{75} & \textbf{74.3} \\

    \ours{LiDSeg (PIDNet-L)} &
      \textbf{99} & \textbf{84} & 89 & 57 & 53 & \textbf{62} & \textbf{66} & 74 & \textbf{92} & 66 & \textbf{96} & \textbf{78} & 58 & \textbf{94} & 67 & 75 & 70 & \textbf{57} & 73 & 74.2 \\

    \ours{LiDSeg (TopFormer-B)} &
      97 & 82 & 89 & 57 & 54 & \textbf{62} & \textbf{66} & \textbf{75} & \textbf{92} & 65 & 95 & 74 & 59 & 92 & 66 & \textbf{76} & 69 & 54 & 74 & 73.6 \\

    \ours{LiDSeg (SeaFormer-B)} &
      97 & 82 & 89 & 56 & 53 & 61 & \textbf{66} & 74 & 91 & 66 & 95 & 73 & 59 & 92 & 65 & \textbf{76} & 70 & 54 & 74 & 73.3 \\
    \bottomrule
  \end{tabular}}

\end{table*}

\begin{table}[t]
  \centering
  \caption{Reconstruction quality vs.\ fog density (BiSeNetV2). Input PSNR
  is foggy vs.\ clear.}
  \label{tab:supp_recon_quality}
  \footnotesize
  \setlength{\tabcolsep}{5pt}
  \begin{tabular*}{\columnwidth}{@{\extracolsep{\fill}}lcccc@{}}
    \toprule
    $\beta$  & \textbf{PSNR} & \textbf{SSIM} & \textbf{mIoU} \\
    \midrule
    0.005 & 34.2 & 0.946 & 76.3 \\
    0.01  & 33.3 & 0.929 & 75.7 \\
    0.02  & 31.0 & 0.902 & 74.2 \\
    \bottomrule
  \end{tabular*}
\end{table}

Beyond synthetic fog analysis, LiDSeg proves to be a robust alternative to fundamental paradigm designs, fully capable of real-world deployment. For instance, pipelining an explicit dehazing model before segmentation (the restore-then-segment paradigm) results in heavy latency and actually underperforms on mIoU (Tab.~\ref{tab:supp_restore}).

\begin{table}[htbp]
  \centering
  \caption{LiDSeg vs.\ restore-then-segment with the same BiSeNetV2 head.
  GFLOPs/FPS at $512{\times}1024$ (Jetson AGX Orin); mIoU at dense fog
  ($\beta=0.02$).}
  \label{tab:supp_restore}
  \footnotesize
  \setlength{\tabcolsep}{5pt}
  \begin{tabular*}{\columnwidth}{@{\extracolsep{\fill}}lcccc@{}}
    \toprule
    \textbf{Pipeline} & \textbf{Params} & \textbf{GFLOPs} &
      \textbf{FPS} & \textbf{mIoU} \\
    \midrule
    DehazeFormer\,$\to$\,BiSeNetV2 & \phantom{0}5.9M & 61.4 & \phantom{0}54 & 71.5 \\
    Restormer\,$\to$\,BiSeNetV2    & 29.4M & 213.0 & \phantom{0}18 & 72.3 \\
    \rowcolor{lightgray}
    \ours{LiDSeg (BiSeNetV2)}      & \textbf{5.22M} & \textbf{27.2} & \textbf{83.4} & \textbf{74.2} \\
    \bottomrule
  \end{tabular*}
\end{table}

\begin{table}[htbp]
  \centering
  \caption{Zero-shot mIoU (\%) on Foggy Zurich~v2 (FZ) and Foggy Driving
  (FD): unguided Semantic segmentation model vs.\ LiDSeg (no target labels).}
  \label{tab:supp_zeroshot}
  \footnotesize
  \setlength{\tabcolsep}{4pt}
  \begin{tabular*}{\columnwidth}{@{\extracolsep{\fill}}lcccc@{}}
    \toprule
    \textbf{Semantic segmentation model} & \textbf{FZ (unguided)} & \textbf{FZ (LiDSeg)} &
      \textbf{FD (unguided)} & \textbf{FD (LiDSeg)} \\
    \midrule
    BiSeNetV2  & 38.2 & \textbf{53.5} & 41.0 & 50.4 \\
    DDRNet-23s & 40.1 & 52.8 & 42.5 & 51.0 \\
    PIDNet-L   & 39.5 & 52.6 & 41.8 & 50.8 \\
    STDC1-Seg  & 39.0 & 53.4 & 42.0 & \textbf{52.4} \\
    \bottomrule
  \end{tabular*}
\end{table}

Most importantly, these robustness improvements do not compromise edge deployability. Complementing the $512{\times}1024$ metrics provided in the main paper, lower-resolution ($256{\times}512$) evaluations confirm that LiDSeg comfortably exceeds real-time thresholds (138--587 FPS) on a Jetson AGX Orin (Tab.~\ref{tab:supp_efficiency_lowres}).

\begin{table*}[htbp]
  \centering
  \caption{Jetson AGX Orin efficiency at $256{\times}512$ (Benchmarking more efficiency in addition to $512{\times}1024$ Tab.~\ref{tab:efficiency}). Guided: LiDSeg
  with the reconstruction decoder discarded at inference.}
  \label{tab:supp_efficiency_lowres}
  \footnotesize
  \setlength{\tabcolsep}{4pt}
  \begin{tabular*}{\textwidth}{@{\extracolsep{\fill}}llrrrr@{}}
    \toprule
    \textbf{Model} & \textbf{Variant} & \textbf{Params\,(M)} & \textbf{GFLOPs} &
      \textbf{FPS} & \textbf{Mem} \\
    \midrule
    BiSeNetV2     & Baseline &  3.36 &  6.23 & 161.5 & 18 \\
    DDRNet        & Baseline &  5.70 &  2.29 & 707.3 & 18 \\
    STDC          & Baseline &  8.43 & 10.39 & 724.2 & 23 \\
    SCTNet-B      & Baseline & 17.44 &  7.04 & 509.3 & 44 \\
    PIDNet-L      & Baseline & 37.31 & 17.26 & 308.0 & 80 \\
    TopFormer-B   & Baseline &  7.06 &  2.42 & 715.2 & 24 \\
    SeaFormer-B   & Baseline &  8.58 &  0.89 & 252.2 & 27 \\
    \midrule
    \rowcolor{lightgray}
    \ours{LiDSeg (BiSeNetV2)}   & Guided &  5.22 &  6.80 & 138.2 & 21 \\
    \rowcolor{lightgray}
    \ours{LiDSeg (DDRNet)}      & Guided &  7.65 &  2.88 & 587.1 & 23 \\
    \rowcolor{lightgray}
    \ours{LiDSeg (STDC)}        & Guided & 11.48 & 11.02 & 545.1 & 33 \\
    \rowcolor{lightgray}
    \ours{LiDSeg (PIDNet-L)}    & Guided & 39.54 & 17.87 & 249.4 & 88 \\
    \rowcolor{lightgray}
    \ours{LiDSeg (TopFormer-B)} & Guided &  9.01 &  3.01 & 512.8 & 28 \\
    \bottomrule
  \end{tabular*}
\end{table*}
\subsection{ACDC: Expert Cross-Condition Generalization}
\label{supp:acdc_master}
To isolate the value of joint training and expose the channel-coherent structure of weather degradation, Tab.~\ref{tab:supp_cross_condition} reports a full cross-condition evaluation of the four condition-specific \textbf{Expert} LiDSeg models. Each Expert was fine-tuned on a single ACDC split and evaluated on all four. The $4{\times}4$ matrix per guided model shows a pattern consistent with our channel-correlation analysis. Feature transfer within the channel-coherent F$\leftrightarrow$R$\leftrightarrow$S triangle is notably strong. Because fog, rain, and snow share a structurally coherent channel degradation characteristic ($\bar\rho>0.8$), any Expert trained on one of these splits retains a large fraction of its diagonal accuracy on the other two. The best off-diagonal entry per model consistently lies inside this triangle, confirming that a shared encoder generalizes well across weather corruptions.

Similar to previous results, night transfer is bounded by scene priors rather than condition-specific features. Every Fog/Rain/Snow Expert evaluated on Night drops below its own diagonal into a narrow 36--43\% range. However, this is still substantially above what a Cityscapes-only Source model achieves on ACDC-Night ($\sim$12--14\%) and comparable to the zero-shot LiDSeg performance on Foggy-Zurich and Foggy-Driving. The symmetric N$\rightarrow$F/R/S rows convey the same point from the opposite direction: Expert Night transfers acceptably to Fog (46.5--50.4\%) but degrade on Rain and Snow (39.3--43.4\%), because nighttime supervision under-samples the high-illumination feature space that those conditions depend on.  Overall, the Uber joint fine-tuning reported in the main paper demonstrates that pooling channel-coherent supervision across all conditions strengthens the shared encoder simultaneously, learning generalized features that natively benefit even disjointed domains like Night.

\begin{table*}[htbp]
  \centering
  \caption{ACDC \emph{val} mIoU (\%) for condition-specific \textbf{Expert} cross-evaluation. Rows: training condition; columns: test condition. \textit{Diagonal}: same-condition performance. \textbf{Bold} and \underline{underline} indicate the best and second-best off-diagonal values, respectively, for each model.}
  \label{tab:supp_cross_condition}
  \footnotesize
  \setlength{\tabcolsep}{3pt}
  \setlength{\aboverulesep}{1pt}
  \setlength{\belowrulesep}{1pt}
  \renewcommand{\arraystretch}{0.88}
  \begin{tabular*}{\textwidth}{@{\extracolsep{\fill}}llcccc@{}}
    \toprule
    \textbf{Model} & \textbf{Expert} & \textbf{Fog} & \textbf{Night} & \textbf{Rain} & \textbf{Snow} \\
    \midrule
    \multicolumn{6}{@{}l}{\textit{BiSeNetV2 (5.22M, 108.8 GFLOPs)}} \\
    \midrule
    BiSeNetV2 & Fog   & \diag{71.9} & 41.7 & \second{54.4} & 51.5 \\
    BiSeNetV2 & Night & 50.4 & \diag{53.1} & 43.4 & 42.1 \\
    BiSeNetV2 & Rain  & 53.3 & 42.8 & \diag{68.4} & 54.2 \\
    BiSeNetV2 & Snow  & 51.8 & 40.7 & \textbf{57.1} & \diag{67.1} \\
    \midrule
    \multicolumn{6}{@{}l}{\textit{DDRNet-23s (7.65M, 46.0 GFLOPs)}} \\
    \midrule
    DDRNet-23s & Fog   & \diag{71.0} & 37.2 & 52.7 & 50.3 \\
    DDRNet-23s & Night & 47.8 & \diag{42.9} & 42.0 & 41.0 \\
    DDRNet-23s & Rain  & 50.6 & 38.0 & \diag{66.9} & \second{53.0} \\
    DDRNet-23s & Snow  & 49.2 & 36.3 & \textbf{55.4} & \diag{66.2} \\
    \midrule
    \multicolumn{6}{@{}l}{\textit{STDC1-Seg (11.48M, 176.3 GFLOPs)}} \\
    \midrule
    STDC1-Seg & Fog   & \diag{71.8} & 38.0 & 51.8 & 52.7 \\
    STDC1-Seg & Night & 48.8 & \diag{44.3} & 41.3 & 43.2 \\
    STDC1-Seg & Rain  & 51.7 & 38.9 & \diag{65.8} & \textbf{55.4} \\
    STDC1-Seg & Snow  & 50.3 & 37.1 & \second{54.5} & \diag{68.0} \\
    \midrule
    \multicolumn{6}{@{}l}{\textit{PIDNet-L (39.54M, 285.9 GFLOPs)}} \\
    \midrule
    PIDNet-L & Fog   & \diag{69.8} & 40.3 & \second{51.7} & 46.6 \\
    PIDNet-L & Night & 46.5 & \diag{51.5} & 41.3 & 39.3 \\
    PIDNet-L & Rain  & 49.3 & 41.3 & \diag{66.9} & 49.1 \\
    PIDNet-L & Snow  & 47.9 & 39.3 & \textbf{54.4} & \diag{63.3} \\
    \bottomrule
  \end{tabular*}
\end{table*}
\subsection{Qualitative Results}
\label{supp:qual}

\begin{figure*}[htbp]
  \centering
  \includegraphics[width=\textwidth]{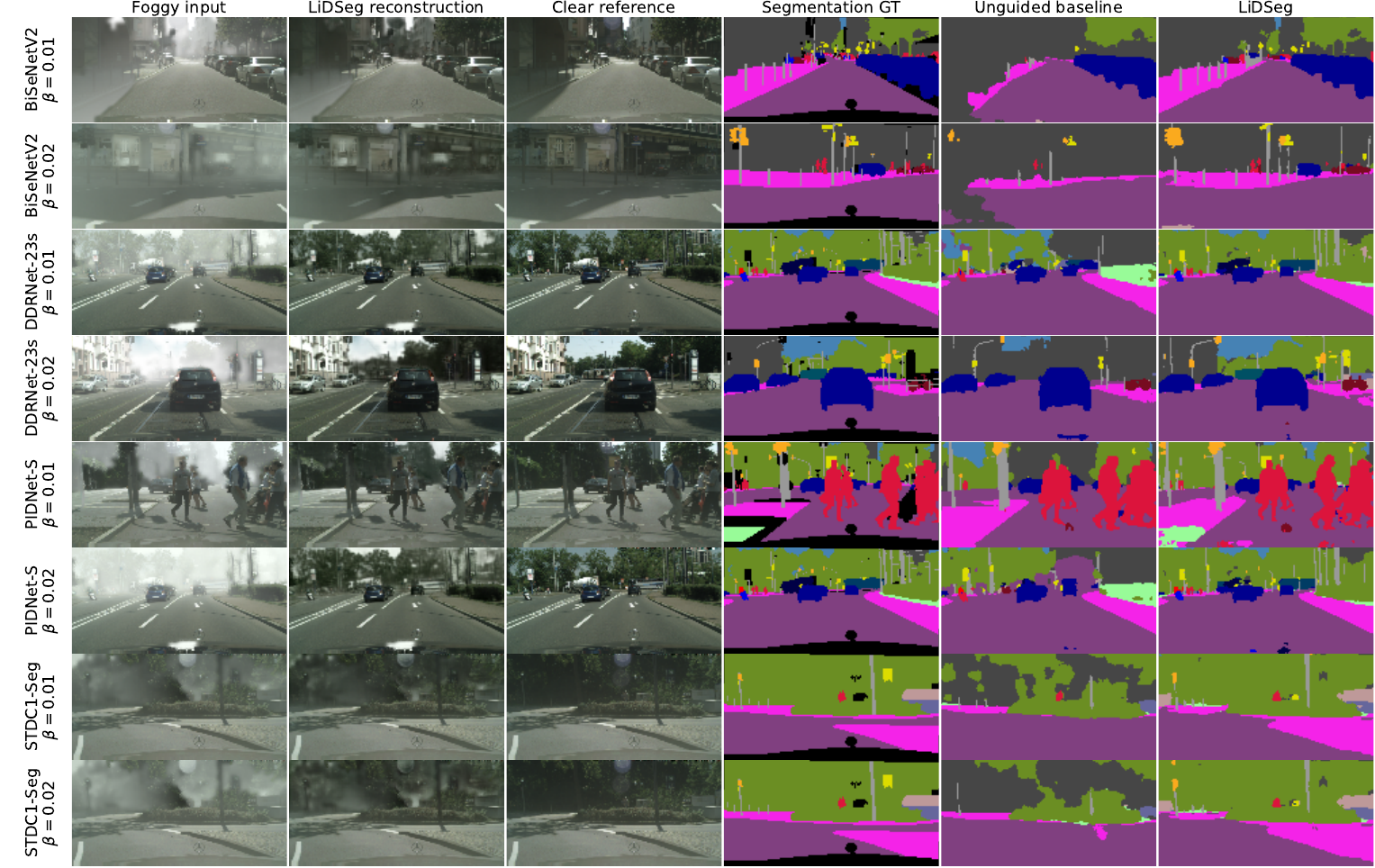}
  \caption{\textbf{Foggy Cityscapes \emph{val}: reconstruction and
  segmentation.} Four Semantic segmentation models at medium ($\beta=0.01$) and dense
  ($\beta=0.02$) fog. Columns: foggy input, LiDSeg reconstruction (training
  only), clear reference, ground truth, unguided baseline, LiDSeg.}
  \label{fig:supp_recon}
\end{figure*}

\begin{figure*}[htbp]
  \centering
  \includegraphics[width=0.8\textwidth]{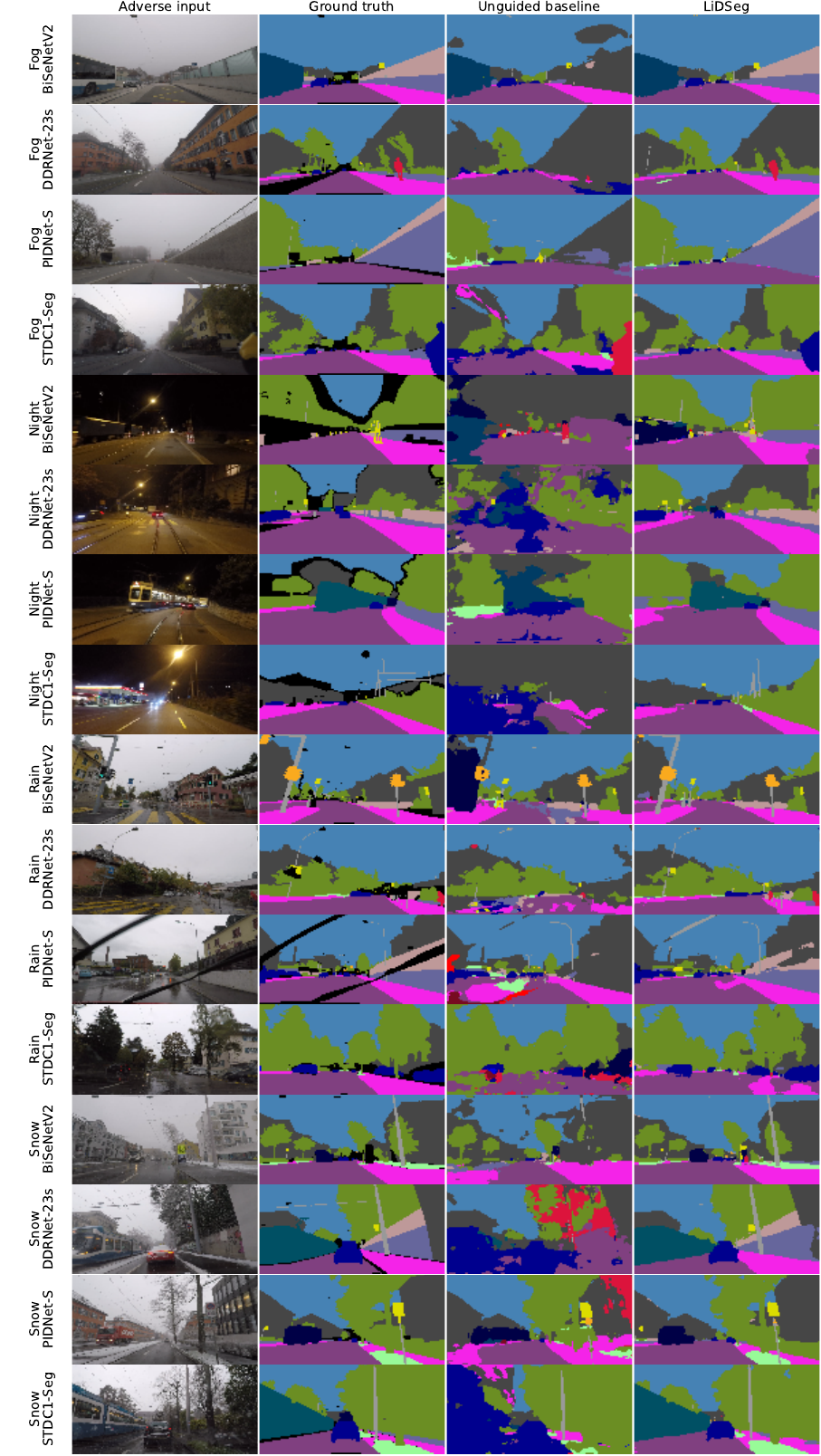}
  \caption{\textbf{Real ACDC \emph{val}: segmentation, all conditions
  for all Semantic segmentation models.} Sixteen distinct scenes (Fog, Night, Rain, Snow
  and four Semantic segmentation models). Columns: adverse input, ground truth, unguided
  baseline, LiDSeg.}
  \label{fig:supp_acdc_recon}
\end{figure*}

Finally, visual comparisons ground our quantitative findings. Fig.~\ref{fig:supp_recon} and Fig.~\ref{fig:supp_acdc_recon} provide qualitative evidence on both synthetic Foggy Cityscapes and real-world ACDC. Foggy Cityscapes includes the training-only reconstruction output alongside segmentation, while ACDC shows segmentation outputs only because no reconstruction decoder is used during ACDC fine-tuning or inference. The examples illustrate preservation of details such as lane markings, pedestrians, and distant objects under adverse conditions.

